\documentclass[10pt,journal,cspaper,compsoc]{IEEEtran}

\ifCLASSINFOpdf

\else

\fi

\usepackage{hyperref}
\usepackage{subfigure}
\usepackage{makeidx}  
\usepackage{graphicx}
\usepackage{amsmath}
\usepackage{nccmath}
\usepackage{extarrows}
\usepackage{amsfonts,amsmath}
\usepackage{algorithm2e}
\usepackage{changepage}
\usepackage{float}

\usepackage{ragged2e}
\usepackage{cite}

\begin{document}

\title{Re-weighting and 1-Point RANSAC-Based\\ P$n$P Solution to Handle Outliers}

\author{Haoyin Zhou,
        Tao Zhang,~\IEEEmembership{Senior Member,~IEEE,}
        and Jayender Jagadeesan,~\IEEEmembership{Member,~IEEE}
\IEEEcompsocitemizethanks{\IEEEcompsocthanksitem
 Haoyin Zhou and Jayender Jagadeesan are with the Surgical Planning Laboratory, Brigham and Women's Hospital, Harvard Medical School, Boston,
 MA, 02115, USA.\protect\\
E-mail: zhouhaoyin@bwh.harvard.edu; jayender@bwh.harvard.edu.
\IEEEcompsocthanksitem Tao Zhang is with the Department
of Automation, School of Information Science and Technology, Tsinghua University, Beijing,
China, 100086.\protect\\
E-mail: taozhang@tsinghua.edu.cn}

\thanks{}}

\markboth{}%
{Shell \MakeLowercase{\textit{et al.}}: Bare Demo of IEEEtran.cls for Computer Society Journals}

\IEEEcompsoctitleabstractindextext{%
\begin{abstract}
\justifying
The ability to handle outliers is essential for performing the perspective-$n$-point (P$n$P) approach in practical applications, but conventional RANSAC+P3P or P4P methods have high time complexities. We propose a fast P$n$P solution named R1PP$n$P to handle outliers by utilizing a soft re-weighting mechanism and the 1-point RANSAC scheme. We first present a P$n$P algorithm, which serves as the core of R1PP$n$P, for solving the P$n$P problem in outlier-free situations. The core algorithm is an optimal process minimizing an objective function conducted with a random control point. Then, to reduce the impact of outliers, we propose a reprojection error-based re-weighting method and integrate it into the core algorithm. Finally, we employ the 1-point RANSAC scheme to try different control points. Experiments with synthetic and real-world data demonstrate that R1PP$n$P is faster than RANSAC+P3P or P4P methods especially when the percentage of outliers is large, and is accurate. Besides, comparisons with outlier-free synthetic data show that R1PP$n$P is among the most accurate and fast P$n$P solutions, which usually serve as the final refinement step of RANSAC+P3P or P4P. Compared with REPP$n$P, which is the state-of-the-art P$n$P algorithm with an explicit outliers-handling mechanism, R1PP$n$P is slower but does not suffer from the percentage of outliers limitation as REPP$n$P.

\end{abstract}

\begin{IEEEkeywords}
Perspective-$n$-Point; 1-Point RANSAC; soft re-weighting; robustness to outliers.
\end{IEEEkeywords}}

\maketitle

\IEEEdisplaynotcompsoctitleabstractindextext

\IEEEpeerreviewmaketitle

\section{Introduction}

\IEEEPARstart{T}{he} perspective-$n$-point (P$n$P) problem aims to determine the position and orientation of a calibrated camera from $n$ known correspondences between three-dimensional (3D) object points and their two-dimensional (2D) image projections. P$n$P is a core problem in the computer vision field and has found many applications, such as robot vision navigation \cite{mur2015orb}, augmented reality \cite{muller2013mobile}, and computer animation. In the past decades, many effective P$n$P approaches have been proposed with very fast computational speed \cite{ferraz2014very}\cite{lepetit2009epnp} and high accuracy \cite{zheng2013revisiting}\cite{lu2000fast}.

To date, most P$n$P algorithms are designed under the assumption that no outlier exists among the given 3D-2D correspondences. However, in practical applications, this outlier-free assumption is often difficult to satisfy. This is because image feature detection and matching approaches, such as SURF \cite{bay2006surf}, BRISK \cite{leutenegger2011brisk} and ORB \cite{rublee2011orb}, do not always give perfect results due to scaling, illumination, shadow and occlusion. Outliers are often unavoidable and they have a significant impact on the P$n$P methods. Even a small percentage of outliers will lead to a significant decrease in accuracy. Hence, the ability to handle outliers is essential for performing P$n$P algorithms in practical applications. The most common outliers handling mechanism is to combine a P$n$P ($n=$ 3 or 4) algorithm \cite{nister2004minimal}\cite{gao2003complete}\cite{josephson2009pose} with the RANSAC-based scheme \cite{fischler1981random} to eliminate outliers, and then perform a more accurate P$n$P algorithm with the remaining inliers to refine the result. A number of very fast closed-form P3P \cite{kneip2011novel} or P4P \cite{bujnak2008general} algorithms have been proposed. However, their RANSAC combination scheme still needs many trials until the selected three or four 3D-2D correspondences are all inliers, which results in a high time complexity. Hence, the computational speed decreases significantly as the percentage of outliers increases. To reduce the time complexity, one natural idea is to utilize the P$n$P algorithm with a smaller $n$. However, when $n=$ 1 or 2, the P$n$P problem has infinitely many solutions, which makes the conventional RANSAC-based scheme infeasible.

To the best of our knowledge, except for RANSAC-based methods, the only P$n$P method that addresses outliers is REPP$n$P \cite{ferraz2014very} proposed by Ferraz \textit{et al}., which is the state-of-the-art P$n$P method that is robust to outliers. REPP$n$P integrates an outlier rejection mechanism with camera pose estimation. It formulates the pose estimation problem as a low-rank homogeneous system in which the solution lies on its one-dimensional (1D) null space. Outlier correspondences are those rows of the linear system that perturb the null space and are progressively detected by projecting them on an iteratively estimated solution of the null space. Although REPP$n$P is very fast and accurate, it suffers from a severe limitation that it cannot handle more than approximately $50\%$ of outliers.

In this paper, we propose a robust 1-point RANSAC-based P$n$P method named R1PP$n$P. We first present an optimal iterative process as the core P$n$P algorithm of R1PP$n$P. The core algorithm takes a random 3D-2D correspondence as the control point. To address outliers, we propose a soft weight assignment method according to reprojection errors to distinguish inliers and outliers, and integrate it into the core algorithm. The weight factors associated with outliers decrease significantly during the iteration to reduce the impact of outliers. Finally, we employ the 1-point RANSAC scheme to try different control points for the core P$n$P algorithm. By using this combination of the RANSAC scheme and the soft weight assignment, the algorithm is capable of eliminating outliers when the selected control point is an inlier.

The main advantage of R1PP$n$P is that it has much lower time complexity and is much faster than conventional RANSAC+P3P or P4P methods, especially when the percentage of outliers is large. Compared with REPP$n$P, the proposed R1PP$n$P does not suffer from the percentage of outliers limitation.

This paper is organized as follows. In Section II, we describe the fundamental model used in R1PP$n$P. The details of the core algorithm are given in Section III, in which we also provide its proof of convergence, local minima analysis and the strategy to select control points. The outliers handling mechanism, including soft weight assignment and the 1-point RANSAC scheme, is introduced in Section IV. We also provide details of termination conditions in Section IV.  Evaluation results are presented in Section V. A discussion and description of planned future work is described in Section VI.

\subsection{Related Works}

The P$n$P problem, coined by Fischler and Bolles \cite{fischler1981random}, is articulated as follows: \emph{Given the relative spatial locations of $n$ control points, and given the angle to every pair of control points $P_i$ from an additional point called the center of perspective $C$, find the lengths of the line segments joining $C$ to each of the control points.} The P$n$P problem has been studied for many years. In early studies, direct linear transformation (DLT) \cite{hartley2003multiple} was used as a solution in a straightforward way by solving a linear system. However, DLT ignores the intrinsic camera parameters, which are assumed to be known, and therefore generally leads to a less stable pose estimate.

In the past decade, researchers have proposed many P$n$P methods to improve speed, accuracy, and robustness to outliers. The P$n$P methods can be roughly classified into non-iterative and iterative methods. Generally speaking, non-iterative methods are more efficient but are unstable under image noise and outliers. Many non-iterative P$n$P methods are based on a set of small number of points ($n=3,4$). They are referred to as P3P \cite{dementhon1992exact} \cite{gao2003complete} \cite{kneip2011novel} or P4P \cite{abidi1995new} \cite{bujnak2008general}\cite{horaud1989analytic}\cite{quan1999linear} methods. P3P is the smallest subset of control
points that yields a finite number of solutions \cite{kneip2011novel}\cite{wolfe1991perspective}. When the intrinsic camera parameters are known and we have $n\geq 4$ points, the solution is generally unique. Triggs proposed a P$n$P method with four- or five- correspondences \cite{triggs1999camera}. These P$n$P methods based on less than four correspondences do not make use of redundant points and are very sensitive to noise and outliers. However, due to their efficiency and capability to calculate from a small point set, P3P or P4P methods are very useful for combing a RANSAC-like scheme to reject outliers. There are also many non-iterative P$n$P methods that are able to make use of redundant points but are quite time consuming. For example, Ansar's method is $O(n^8)$ \cite{ansar2003linear} and Fiore's is $O(n^2)$ \cite{fiore2001efficient}. Schweighofer proposed an $O(n)$ P$n$P method named SDP, but is slow \cite{schweighofer2008globally}. In recent years, three excellent $O(n)$ effective non-iterative P$n$P methods, EP$n$P \cite{lepetit2009epnp}, RP$n$P \cite{li2012robust} and UP$n$P\cite{kneip2014upnp}, have been proposed, and these methods are very efficient and accurate even compared to iterative methods.

Iterative P$n$P methods \cite{lu2000fast}\cite{lowe1991fitting}\cite{dementhon1995model}\cite{horaud1997object} are mostly optimization methods that decrease their energy function in the iterative process. They are generally more accurate and robust, but slower. For example, Dementhon proposed POSIT that is easy to implement \cite{dementhon1995model} and further proposed SoftPOSIT to handle situations when the correspondence relationships are unknown \cite{david2004softposit}. Although SoftPOSIT has a certain ability to handle outliers, the strong assumption that all correspondences are unknown make it slow. Lu's method \cite{lu2000fast} is the most accurate iterative P$n$P method but may get stuck in local minima. Schweighofer discussed the local minima situation of Lu's method and proposed a method to avoid this limitation \cite{schweighofer2006robust}.

P$n$P algorithms are widely used in applications such as structure from motion \cite{havlena2010efficient} and monocular SLAM \cite{mur2014fast}, which require dealing with hundreds or even thousands of noisy feature points and outliers in real-time. The fact that outliers have a much greater impact on P$n$P accuracy than image Gaussian white noise makes it is necessary for the P$n$P algorithm to handle outliers efficiently. Conventional method to handle outliers is to combine a RANSAC-like scheme with the P3P or P4P algorithms. Besides, L1-norm is also widely used to handle a certain amount of outliers \cite{kahl2008practical}\cite{ke2007quasiconvex} because the L1-norm penalty is less sensitive to outliers than the L2-norm penalty. Although a L1-norm-based energy function is more robust to outliers, it cannot absolutely get rid of outliers and its computation is more complex.

Ferraz \textit{et al}. proposed a very fast P$n$P method that can handle up to 50\% of outliers \cite{ferraz2014very}. The outlier rejection mechanism is integrated within the pose estimation pipeline with negligible computational overhead. Compared to Ferraz's method, the R1PP$n$P algorithm proposed in this paper demonstrates much stronger robustness, but is slower.

\section{Fundamental Model}

In this paper we denote the camera frame as $c$ and the world frame as $w$. For point $i$, without taking into account the distortion, the perspective projection equations are employed to describe the pinhole camera model,

\begin{equation}
{u_i} = f\frac{{x_i^c}}{{z_i^c}},{v_i} = f\frac{{y_i^c}}{{z_i^c}},
\label{eq_1}
\end{equation}

\noindent where $f$ is the camera focal length, ${\bf{x}}_i = {[{u_i},{v_i},f]^T}$ is the image homogeneous coordinate in pixels, and ${\bf{X}}_i^c = [x_i^c,y_i^c,z_i^c]^T$ is the real-world coordinate with respect to the camera frame.

According to \eqref{eq_1},

\begin{equation}
{\bf{X}}_i^c = \lambda_i^* {\bf{x}}_i,
\label{eq_2}
\end{equation}

\noindent where $\lambda_i^* = z_i^c / f$ is the normalized depth of point $i$. \eqref{eq_2} indicates that an object point lies on the straight line of sight of the related image point.

The relationship between the camera and world frame coordinate of point $i$ is
\begin{equation}
{\bf{X}}_i^c = {\bf{R}}{\bf{X}}_i^w + {\bf{t}},
\label{eq_3}
\end{equation}

\noindent where ${\bf{R}}\in SO(3)$ is the rotation matrix and ${\bf{t}}\in R^3$ is the translation vector. ${\bf{R}}$ and ${\bf{t}}$ are the variables that need to be estimated in the P$n$P problem.

Similarly to the translation elimination method used in works \cite{kneip2013using}\cite{lee2015minimal}, with two points $i$ and $o$,

\begin{equation}
{\bf{X}}_i^c - {\bf{X}}_{o}^c = {\bf{R}}({\bf{X}}_i^w - {\bf{X}}_o^w),{\kern 1pt} {\kern 1pt} {\kern 1pt} {\kern 1pt} {\kern 1pt} {\kern 1pt} i \ne o.
\label{eq_4}
\end{equation}

In the proposed R1PP$n$P algorithm, $o \in [1,N]$ suggests the index of the control point, $N$ is the number of 3D-2D correspondences. R1PP$n$P represents the shape of the point cloud by the relative positions between the control point $o$ and other $N-1$ points. Denoting ${\bf{S}}_i = {\bf{X}}_i^w - {\bf{X}}_o^w$, where $S$ means "shape", then, according to \eqref{eq_2} and \eqref{eq_4},

\begin{equation}
\lambda_i^* {\bf{x}}_i - \lambda_o^* {\bf{x}}_o = {\bf{R}}{\bf{S}}_i.
\label{eq_5}
\end{equation}

We divide both sides of \eqref{eq_5} by the depth of the control point $\lambda_o^*$, and rewrite \eqref{eq_5} as

\begin{equation}
\lambda_i {\bf{x}}_i - {\bf{x}}_o = \mu{\bf{R}}{\bf{S}}_i,
\label{eq_6}
\end{equation}

\noindent where $\lambda_i = \mu \lambda_i^*$ and $\mu=1/\lambda_o^*$ is the scale factor. We have

\begin{equation}
{\bf{t}} =  1/\mu {\bf{x}}_o - {\bf{RX}}_o^w.
\label{eq_7}
\end{equation}


According to \eqref{eq_6} and \eqref{eq_7}, the P$n$P problem can be solved by minimizing the objective function

\begin{equation}
f({\bf{R}},\mu,{\lambda _i}) = \sum\limits_{i = 1,i\neq o}^N {{{\left\| {{\lambda _i}{{\bf{x}}_i} - {{\bf{x}}_o} - \mu {\bf{R}}{{\bf{S}}_i}} \right\|^2}}},
\label{eq_objfun1}
\end{equation}

\noindent where $\left\|  \cdot  \right\|$ is the L2-norm.

The objective function \eqref{eq_objfun1} is based on Euclidean distances in the 3D space. Compared with the reprojection error cost, Eq.\eqref{eq_objfun1} gives more weights to points with larger depths. For example, the same level of reprojection error has relatively larger effects when related to an object point with greater depth. To solve this problem, we normalize the cost function \eqref{eq_objfun1} with depths of points and propose the objective function of our R1PP$n$P algorithm, that is

\begin{equation}
f({\bf{R}},\mu ,{\lambda _i}) = {\sum\limits_{i = 1,i \ne o}^N {\left( {\frac{1}{{{\lambda _i}}}\left\| {{\lambda _i}{{\bf{x}}_i} - {{\bf{x}}_o} - \mu {\bf{R}}{{\bf{S}}_i}} \right\|} \right)} ^2},
\label{eq_objfun1_lambda}
\end{equation}

\noindent where $1/\lambda_i$ is introduced to adjust the weight of point $i$ to eliminate the inequity among points in Eq.\eqref{eq_objfun1}.

We estimate $\bf{R}$, $\mu$ and $\lambda_i$ $(i=1,...,N,i\neq o)$ by minimizing the objective function \eqref{eq_objfun1_lambda}, the variables of which consist of two parts: the camera pose \{$\bf{R}$, $\mu$\} and the relative depths with respect to the control point $\{ \lambda_i \}$. To describe the following algorithm intuitively, we introduce two sets of points: ${\bf{p}}_i$ and ${\bf{q}}_i$. With a randomly selected control $o$, points ${\bf{p}}_i$ are determined by the camera pose \{$\bf{R}$, $\mu$\}, and points ${\bf{q}}_i$ are determined by depths $\lambda_i$. We have

\begin{equation}
{\bf{p}}_i = {\bf{x}}_o + \mu {\bf{RS}}_i,
\label{eq_p}
\end{equation}

\begin{equation}
{\bf{q}}_i = \lambda_i {\bf{x}}_i.
\label{eq_q}
\end{equation}

\begin{figure} [ht]
\vspace{0.0cm}
\centering
  \includegraphics[height =8.5cm]{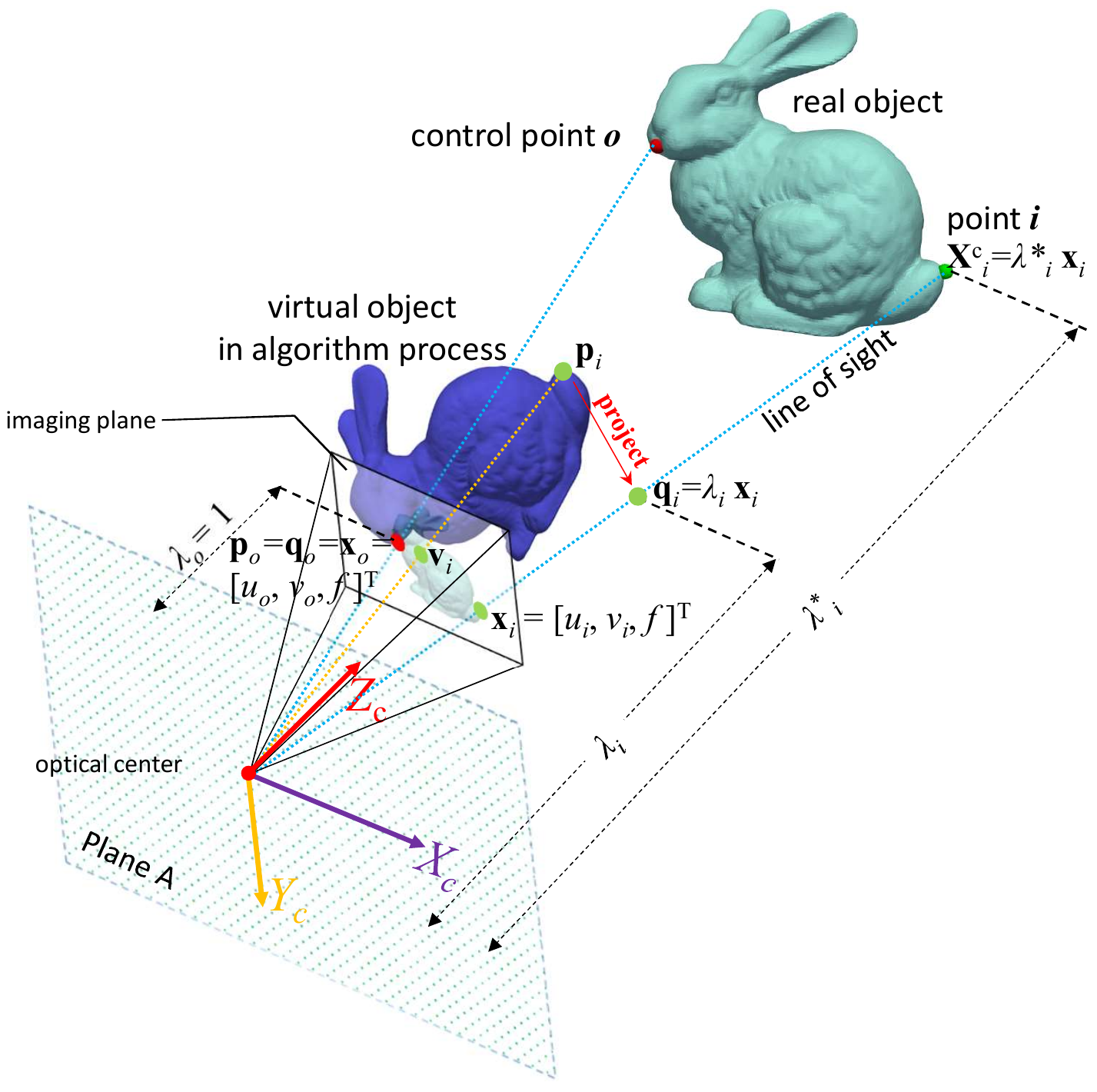}
 \caption{Demonstration of geometrical relationships with a bunny model. The mouth point is the control point $o$. In algorithm, all virtual points rotate and scale around the control point ${\bf{p}}_o={\bf{q}}_o={\bf{x}}_o$. We use the tail point to exemplify ${\bf{p}}_i$ and its projection ${\bf{q}}_i$. Plane A is parallel to the imaging plane and passes the camera optical center. Without loss of generality and for clearer demonstration, in this figure we use focal length $f=1$ and all depths are distances between points and plane A.}
\label{fig_bunnyimg}
\end{figure}

As shown in Fig. \ref{fig_bunnyimg}, points ${\bf{p}}_i$ are attached with the virtual object obtained by rotating and scaling the real object around the control point ${\bf{p}}_o = {\bf{q}}_o = {\bf{x}}_o$. ${\bf{q}}_i$ is the projection of ${\bf{p}}_i$ on the corresponding line of sight. The objective function \eqref{eq_objfun1_lambda} is equivalent to

\begin{equation}
f({{\bf{p}}_i},{{\bf{q}}_i}) = {\sum\limits_{i = 1,i \ne o}^N {\left( {\frac{1}{{{\lambda _i}}}\left\| {{{\bf{p}}_i} - {{\bf{q}}_i}} \right\|} \right)} ^2}.
\label{eq_objfun2}
\end{equation}

As this objective function approaches the global optimal solution and as shown in Fig. \ref{fig_bunnyimg}, point ${\bf{p}}_i$ gets close to ${\bf{q}}_i$ and the $z$-component of ${\bf{p}}_i$ gets close to $f\lambda_i$. Hence, it is expected that the objective function \eqref{eq_objfun2} has similar optimal solutions as the conventional reprojection error cost, because

\begin{equation}
\begin{array}{*{20}{l}}
{f({{\bf{p}}_i},{{\bf{q}}_i})}\\
{ = \sum\limits_{i = 1,i \ne o}^N {{{\left\| {{{\left[ {\frac{{{{\bf{p}}_{i,x}}}}{{{\lambda _i}}},\frac{{{{\bf{p}}_{i,y}}}}{{{\lambda _i}}},\frac{{{{\bf{p}}_{i,z}}}}{{{\lambda _i}}}} \right]}^T} - {{\bf{x}}_i}} \right\|}^2}} }\\
{ \approx \sum\limits_{i = 1,i \ne o}^N {{{\left\| {{{\left[ {f\frac{{{{\bf{p}}_{i,x}}}}{{{{\bf{p}}_{i,z}}}},f\frac{{{{\bf{p}}_{i,y}}}}{{{{\bf{p}}_{i,z}}}},f} \right]}^T} - {{\left[ {{u_i},{v_i},f} \right]}^T}} \right\|}^2}} }.
\end{array}
\label{eq_objfun2_explaination}
\end{equation}

\section{Core Algorithm Design}

We first introduce the core algorithm of R1PP$n$P, which solves the P$n$P problem in outlier-free situations. This section introduces the core algorithm process, proof of convergence, the local minima avoidance mechanism and the strategy to select the control point.

The core algorithm of R1PP$n$P is an optimal iterative process with the objective function \eqref{eq_objfun1_lambda} or \eqref{eq_objfun2}. In each iteration, it estimates the points set ${\bf{q}}_i$ and ${\bf{p}}_i$ $(i=1,...,N,i\neq o)$ alternately by fixing one points set and updating the other one according to the objective function minimization.

\noindent \textbf{(1) ${\bf{q}}_i$ estimation stage.}

Because each ${\bf{q}}_i$ are independent with each other, our algorithm seeks the closest ${\bf{q}}_i$ for each ${\bf{p}}_i$. According to \eqref{eq_q}, points ${\bf{q}}_i$ are constrained to the related lines of sight. Hence, we vertically project ${\bf{p}}_i$ onto the related lines of sight to obtain the points' relative depths with respect to the control point $o$ by

\begin{equation}
\lambda_i =  {\bf{x}}_i^T {\bf{p}}_i / ({\bf{x}}_i^T {\bf{x}}_i),  {\kern 5pt} i = 1,...,N {\kern 5pt} {\rm{and}} {\kern 5pt} i\ne o.
\label{eq_lambda}
\end{equation}

Then, points ${\bf{q}}_i$ are updated according to Eq.\eqref{eq_q}.

\noindent \textbf{(2) ${\bf{p}}_i$ estimation stage.}

Points ${\bf{p}}_i$ are determined by $\bf{R}$ and $\mu$. According to \eqref{eq_p}, the updated $\bf{R}$ and $\mu$ should make points \{$\mu {\bf{RS}}_i$\} have the smallest weighted sum of squared distances to points \{${\bf{q}}_i - {\bf{x}}_o$\}, and subject to ${\bf{R}}^T{\bf{R}} = {\bf{I}}_{3\times 3}$.  According to the objective function \eqref{eq_objfun2}, the weights used in this stage are $1/\lambda_i$ in the previous iteration.

Denoting matrices ${\bf{A}} = \left[ {\frac{{{{\bf{q}}_1} - {{\bf{x}}_o}}}{{{\lambda _1}}},...,\frac{{{{\bf{q}}_N} - {{\bf{x}}_o}}}{{{\lambda _N}}}} \right]_{3\times N}$ and ${\bf{S}} = \left[ {\frac{{{{\bf{S}}_1}}}{{{\lambda _1}}},...,\frac{{{{\bf{S}}_N}}}{{{\lambda _N}}}} \right]_{3\times N}$, then according to Ref. \cite{arun1987least}
\begin{equation}
[{\bf{U}},{\bf{\Sigma }},{{\bf{V}}^T}] = {\rm{svd}}({\bf{A}}{{\bf{S}}^T}), {\bf{R}} = {\bf{U}}{{\bf{V}}^T},
\label{eq_R}
\end{equation}

\begin{figure} [!htb]
\vspace{0.0cm}
\centering
  \includegraphics[height = 5.2cm]{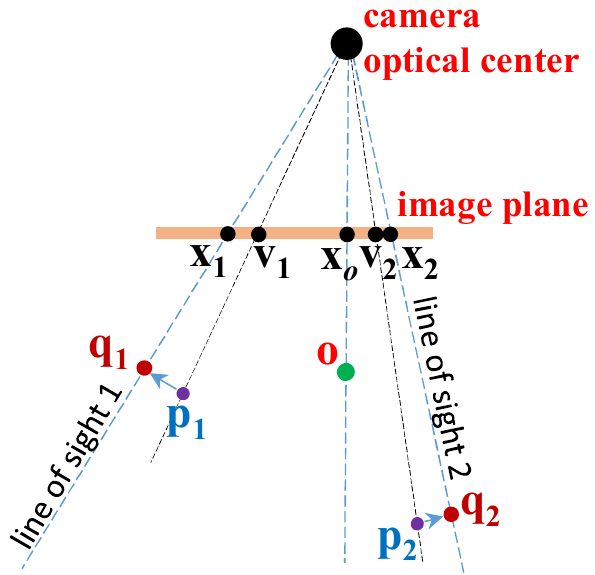}
\caption{Demonstration of the updating method of the scale factor $\mu$. One possible method is to update $\mu$ according to the Euclidean distances between ${\bf{p}}_i$, ${\bf{q}}_i$ and $\bf{o}$, which works for ${\bf{p}}_1$ and ${\bf{q}}_1$ because they have close depths as $\bf{o}$. However, this method may result in slow $\mu$ updating rate for ${\bf{p}}_2$ and ${\bf{q}}_2$ because $\left\| {\bf{q}}_2 - {\bf{o}} \right\| \approx \left\| {\bf{p}}_2 - {\bf{o}} \right\|$. Hence, it is more efficient to compare ${\bf{v}}_i$ and ${\bf{x}}_i$ to move points ${\bf{p}}_i$ to the related lines of sight.}
\label{fig_newmu}
\end{figure}

Because points ${\bf{p}}_i$ are directly generated from ${\bf{S}}_i$ according to Eq.\eqref{eq_p}, Eq.\eqref{eq_R} suggests that $\bf{R}$ is updated according to the differences between points ${\bf{p}}_i$ and ${\bf{q}}_i$ in the 3D space. However, Fig.\ref{fig_newmu} demonstrates that by using this method, the updating rate of $\mu$ may be slow in situations when the range of depths is large. To achieve faster convergence rate, we update the scale factor $\mu$ by comparing the projected image coordinates of ${\bf{p}}_i$, which are denoted as ${{\bf{v}}_i}$, and the real image points ${\bf{x}}_i$. Denoting matrices ${\bf{B}} = \left[ {{{\bf{v}}_1} - {{\bf{x}}_{o}},...,{{\bf{v}}_N} - {{\bf{x}}_o}} \right]_{3\times N}$ and ${\bf{C}} = \left[ {{{\bf{x}}_1} - {{\bf{x}}_o},...,{{\bf{x}}_N} - {{\bf{x}}_o}} \right]_{3\times N}$. $\mu$ is updated by

\begin{equation}
\Delta \mu  = \left\| {{\rm{vector}}({\bf{C}})} \right\|/\left\| {{\rm{vector}}({\bf{B}})} \right\|
\label{eq_mu1}
\end{equation}

\begin{equation}
{\mu _{{\rm{new}}}} = {\mu _{{\rm{old}}}}\Delta \mu
\label{eq_mu2}
\end{equation}

Finally, points ${\bf{p}}_i$ are updated according to Eq.\eqref{eq_p}.

\subsection{Proof of Convergence}

We first provide the mathematical proof of the convergence of R1PP$n$P when not using $1/\lambda_i$ as weights in the objective function \eqref{eq_objfun2}. $k$ denotes the number of iterations, ${\bf{q}}_i^{(k+1)}$ is obtained by vertically projecting ${\bf{p}}_i^{(k)}$ to the line of sight $i$, and ${\bf{q}}_i^{(k+1)}$ and ${\bf{q}}_i^{(k)}$ are on the line of sight $i$. Hence, the three points, ${\bf{p}}_i^{(k)}$,${\bf{q}}_i^{(k)}$, and ${\bf{q}}_i^{(k+1)}$ comprise a right-angled triangle. Therefore, for each index $i,i \neq o$,

\begin{equation}
\left\| {{\bf{p}}_i^{(k)}-{\bf{q}}_i^{(k+1)}} \right\|^2 = \left\| {{\bf{p}}_i^{(k)}-{\bf{q}}_i^{(k)}} \right\|^2 - \left\| {{\bf{q}}_i^{(k + 1)}-{\bf{q}}_i^{(k)}} \right\|^2.
\label{eq_8}
\end{equation}

In the ${\bf{p}}^{(k+1)}$ updating stage, the updated $\bf{R}$ and $\mu$ make the objective function \eqref{eq_objfun2} smaller. Hence,

\begin{equation}
\sum\limits_{i = 1}^N {{{\left\| {{\bf{p}}_i^{(k + 1)} - {\bf{q}}_i^{(k + 1)}} \right\|}^2}}  \le \sum\limits_{i = 1}^N {{{\left\| {{\bf{p}}_i^{(k)} - {\bf{q}}_i^{(k + 1)}} \right\|}^2}} .
\label{eq_9}
\end{equation}

According to \eqref{eq_8}, \eqref{eq_9}, and the objective function \eqref{eq_objfun2},

\begin{equation}
\begin{array}{l}
\begin{array}{l}
f{({\bf{p}}_i^{(k + 1)},{\bf{q}}_i^{(k + 1)})} \le f{({\bf{p}}_i^{(k)},{\bf{q}}_i^{(k)})} - \sum\limits_{i = 1}^N {{{\left\| {{\bf{q}}_i^{(k + 1)} - {\bf{q}}_i^{(k)}} \right\|}^2}}
\end{array}.
\end{array}
\label{eq_10}
\end{equation}

Hence, the objective function will strictly decrease until ${\bf{q}}_i^{(k + 1)} = {\bf{q}}_i^{(k)}$ when not using $1/\lambda_i$ as weights. However, when $1/\lambda_i$ is applied in the objective funtion, the above convergence proof is not rigorous in mathematics because $\lambda_i^{(k+1)} \neq \lambda_i^{(k)}$. As the iteration process, the changes of $\lambda_i$ become small, which makes the formula \eqref{eq_10} hold. In addition, our experimental results in this paper also support the assumption that our algorithm is convergent.

\subsection{Local Minima Avoidance}

\begin{figure} [ht]
\vspace{0.0cm}
\centering
  \includegraphics[height =7.1cm]{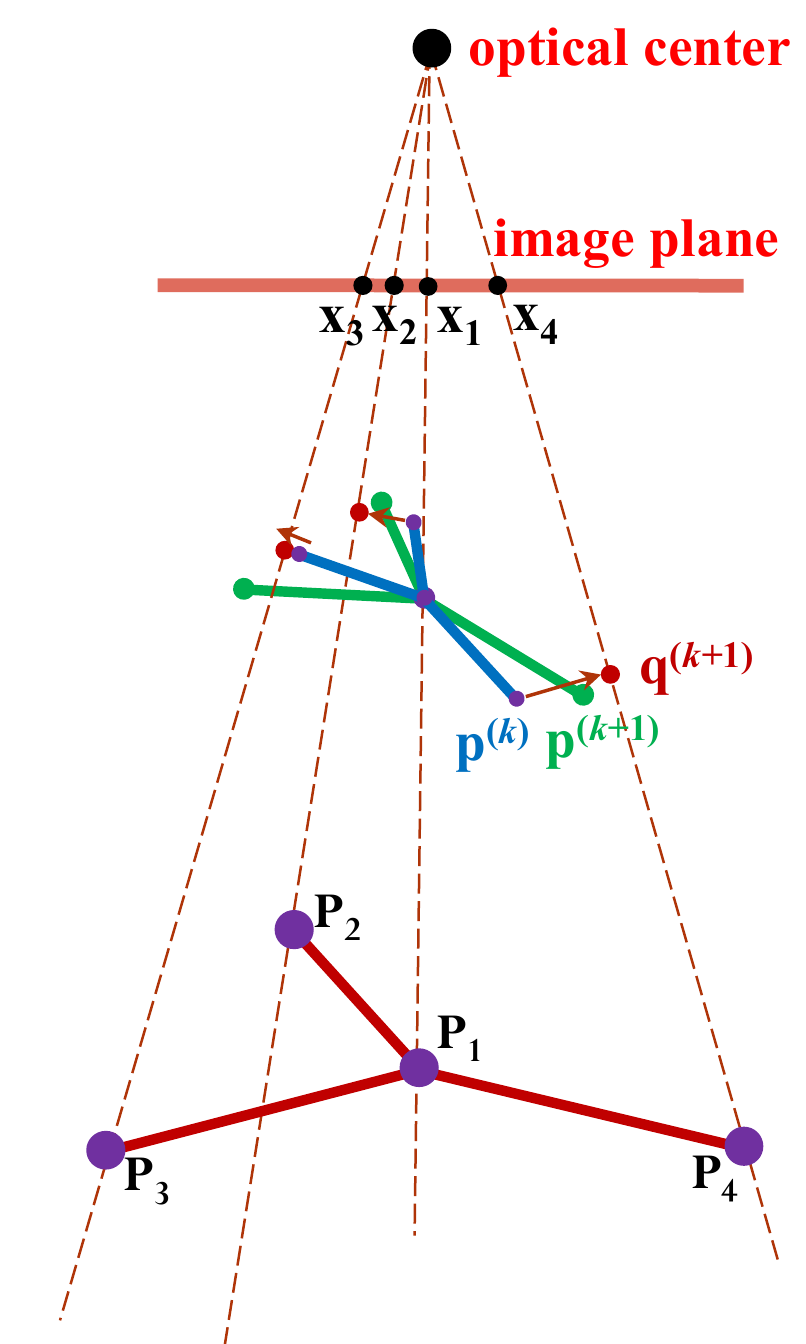}
  \includegraphics[height =7.0cm]{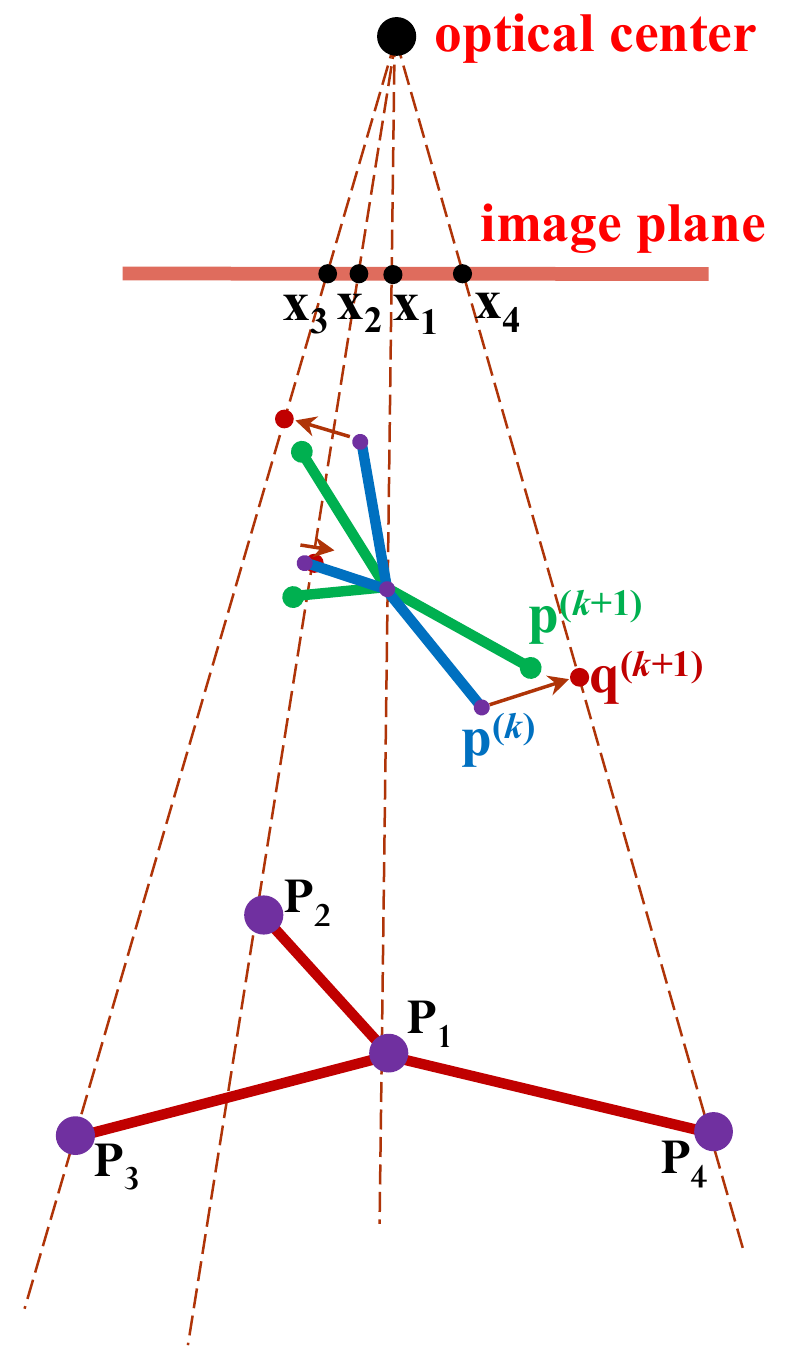}
\caption{Iterative process with 2D space and 1D camera imaging plane. ${\bf{P}}_1$, ${\bf{P}}_2$, ${\bf{P}}_3$ and ${\bf{P}}_4$ are the object points; ${\bf{P}}_1$ is selected as the control point ($o=1$). (a) The process ${\bf{p}}^{(k)} \rightarrow {\bf{q}}^{(k+1)}\rightarrow {\bf{p}}^{(k+1)}\rightarrow ... $ makes the estimation pose approach the correct solution; (b) the rotation related to ${\bf{p}}^{(k+1)}$ is worse than that related to ${\bf{p}}^{(k)}$, which means the process is approaching a local minima, which is a mirror-image form of the true object shape.}
\label{fig_iterprocess}
\end{figure}

We have concluded that the iterative process of R1PP$n$P is convergent. However, we still need to address situations that R1PP$n$P may get stuck in local minima. To demonstrate the iterative process more intuitively, we introduce a 1D camera working in the 2D space, as shown in Fig. \ref{fig_iterprocess}. In this demonstration, an object with four points ${\bf{P}}_i$, $i=1,...,4$ are projected to the camera image plane and their image points ${\bf{x}}_i$ are obtained. ${\bf{P}}_1$ is selected as the control point, which means $o=1$. Different initial values may result in different convergence results.

Fig. \ref{fig_iterprocess}(a) demonstrates a process that is approaching the correct global optimal results. Beginning with points ${\bf{p}}^{(k)}$, the algorithm projects ${\bf{p}}^{(k)}$ to their related lines of sight and obtains points ${\bf{q}}^{(k+1)}$. Then, according to ${\bf{q}}^{(k+1)}$, the algorithm updates the rotation $\bf{R}$ and scale factor $\mu$ to generate points ${\bf{p}}^{(k+1)}$. In this process, the rotation and scale factor related to ${\bf{p}}^{(k+1)}$ are closer to the truth compared to that related to ${\bf{p}}^{(k)}$, and finally the algorithm will reach the correct solution.

However, as the progress shown in Fig. \ref{fig_iterprocess}(b) indicates, ${\bf{p}}^{(k+1)}$ has a larger pose error than ${\bf{p}}^{(k)}$, and the algorithm will finally get stuck in local minima. The reconstructed points ${\bf{p}}^{(k)}$ or ${\bf{p}}^{(k+1)}$ come in mirror-image forms of the real object points ${\bf{P}}$. Without loss of generality, in either 2D or 3D space, a mirror-image form suggests that the left-right-handed shape of the point cloud has been changed, which should not happen in reality. The reason the core algorithm of R1PP$n$P may generate points with different left-right-handed shape is that its rotation estimation equation \eqref{eq_R} does not constrict $\det ({\bf{R}}) = 1$.

In practice we found it not appropriate to constrain $\det ({\bf{R}}) = 1$ from the beginning of the algorithm. Instead, we allow the iteration process to approach the mirror-image form. This is because we found that, with the constrain $\det ({\bf{R}}) = 1$ from the beginning, the algorithm has many types of local minima and they are unpredictable. However, without this constrain, the convergence direction of the core algorithm becomes predictable, with only two types of convergence. The algorithm may reach the global optimal result directly or the approximate mirror-image form. For the latter case, the estimated $\det ({\bf{R}}) = -1$.

Hence, according to the above analysis, we propose the local minima avoidance mechanism. The algorithm begins with a random initial value and control point. When the algorithm converges to a result with $\det ({\bf{R}}) = -1$, we perform a mirror flip by

\begin{equation}
{\lambda _{i,\rm{new}}} = 1/{\lambda _{i,\rm{old}}},{\kern 5pt} i = 1,...,N {\kern 5pt} {\rm{and}} {\kern 5pt} i\ne o.
\label{eq_lambdaflip}
\end{equation}

\subsection{Control Point $o$ Selection}

\begin{figure} [htp]
\vspace{0.0cm}

\centering
 \subfigure[]{
  \includegraphics[height = 3.8cm]{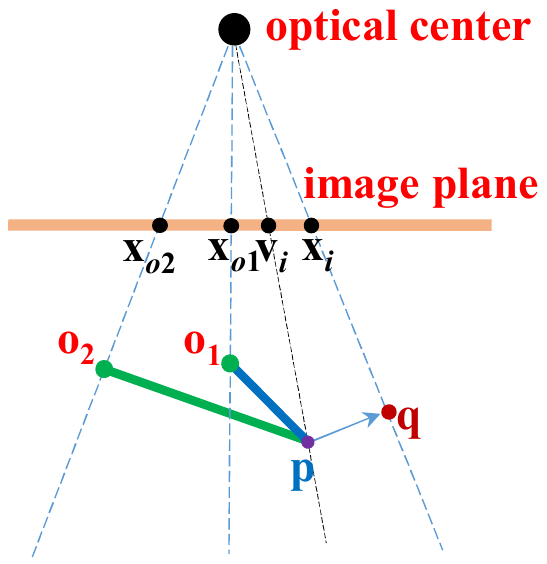}
    } 
 \subfigure[]{
  \includegraphics[height = 3.2cm]{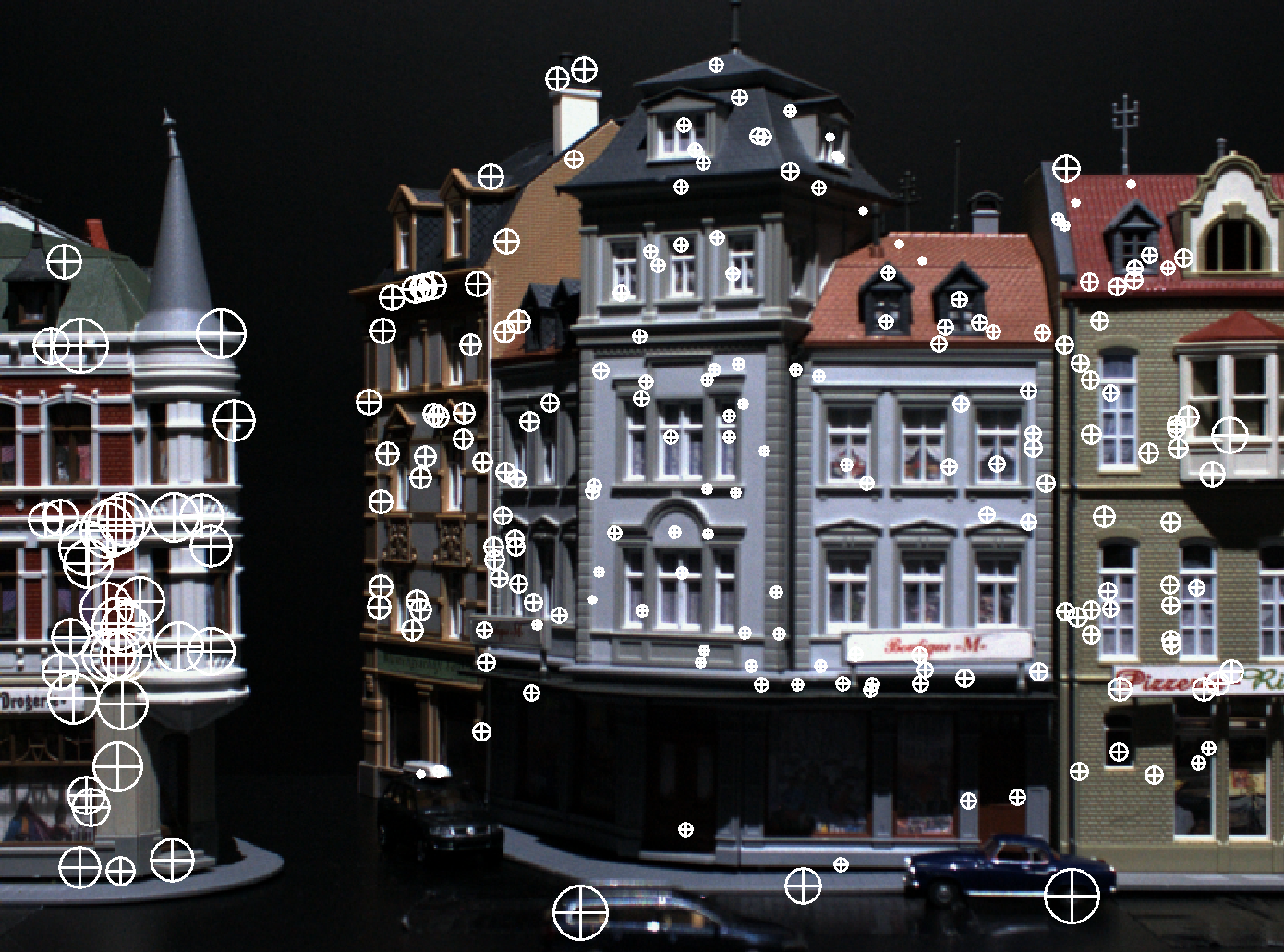}
    }

  \caption{In R1PP$n$P, the selection of control point $o$ is related to the convergence rate. (a) An example to illustrate this behavior with 2D space and 1D camera imaging plane . According to the $\bf{R}$ and $\mu$ updating methods in Eqs.\eqref{eq_R} and \eqref{eq_mu1}. $\angle \left( {{p} - {{{o}}_1},{q} - {{{o}}_1}} \right) > \angle \left( {{p} - {{o}_2},{q} - {{o}_2}} \right)$ and $\left\| {{{\bf{x}}_i} - {{\bf{x}}_{o1}}} \right\|/\left\| {{{\bf{v}}_i} - {{\bf{x}}_{o1}}} \right\| > \left\| {{{\bf{x}}_i} - {{\bf{x}}_{o2}}} \right\|/\left\| {{{\bf{v}}_i} - {{\bf{x}}_{o2}}} \right\|$ suggest that the iteration process is more likely to have larger $\bf{R}$ and $\mu$ updating rate when $o$ is closer to ${p}$. (b) A real-world example to illustrate this behavior, the radius of a circle represents the required number of iterations when using this feature point as the control point $o$.}
\label{fig_convergencerate}
\end{figure}

To select different points as the control point $o$ may result in different convergence rates. Without taking into account noise, the correct value of rotation $\bf{R}$ should be the same for any control point $o$ in a P$n$P task. Hence, larger rotation updating steps in the iteration process suggest that less number of iterations are required to converge to the correct value when starting from the same initial value. In R1PP$n$P, $\bf{R}$ is updated according to the differences between points ${\bf{p}}_i$ and ${\bf{q}}_i$, $i=1,...,N, i \neq o$. When point $o$ is close to ${\bf{p}}_i$, the rotation updating steps are more likely to be large, as shown in Fig.\ref{fig_convergencerate}(a). The updating rates of $\mu$ also follow this analysis. Therefore, we are prone to select the control point $o$ from the center of the point cloud, which has better odds of having smaller distances to the rest of the point cloud to achieve faster convergence rate, as shown in Fig.\ref{fig_convergencerate}(b).

\section{Outliers Handling Mechanism}

The robust and fast capability of handling outliers is the main contribution of the proposed R1PP$n$P algorithm. Our outliers handling mechanism combines a soft weight assignment method and the 1-point RANSAC scheme.

\subsection{Soft Re-weighting}

R1PP$n$P mainly consists of ${\bf{q}}_i$ and ${\bf{p}}_i$ estimation stages. As described in Section 3, in the ${\bf{q}}_i$ stage, calculations related to each point are independent from the others. Hence, outliers do not affect inliers in the ${\bf{q}}_i$ stage. However, in the ${\bf{p}}_i$ stage, outliers perturbs the camera pose estimation results. To reduce the impact of outliers, the basic idea of our soft re-weighting method is to assign each 3D-2D correspondence a weight factor, and to make weight factors related to outliers small when estimating the camera pose in the ${\bf{p}}_i$ stage.

One possible method to assign weights is based on least median of squares \cite{simpson1997introduction}, however this method cannot handle more than $50\%$ of outliers. We designed a soft weight assignment method embedded in the iteration process. To distinguish inliers and outliers, the weights of 3D-2D correspondences are determined by

\begin{equation}
\label{eq_reweighting}
{w_i} = \left\{ {\begin{array}{*{20}{c}}
{1.0}\\
{H/{e_i}}
\end{array}{\kern 1pt} {\kern 1pt} {\kern 1pt} {\kern 1pt} {\kern 1pt} {\kern 1pt} {\kern 1pt} {\kern 1pt} \begin{array}{*{20}{c}}
{if{\kern 1pt} {\kern 1pt} {e_i} \le H}\\
{if{\kern 1pt} {\kern 1pt} {e_i} > H}
\end{array}} \right.,
\end{equation}

\noindent where $e_i$ suggests the reprojection error of point $i$ with the current $\bf{R}$ and $\mu$ during iteration, $H$ is the inliers threshold that points with final reprojection errors smaller than $H$ are considered as inliers. The reweighting rule \eqref{eq_reweighting} suggests that a point with a large reprojection error will have a small weight during the estimation of camera pose, which is designed under a reasonable assumption that outliers have much larger reprojection errors than inliers. Although inliers may also have larger reprojection errors than $H$ during the iteration process, it is acceptable to assign weights that are smaller than 1 to inliers as long as outliers have much smaller weights. Hence, we simply use $H$ as the benchmark to assign weights.

According to $\bf{R}$ estimation given by equations \eqref{eq_R}, we multiply the weight factors with each item of matrices $\bf{A}$ and $\bf{S}$,

\begin{equation}
{\bf{A}}_i = w_i {\bf{A}}_i, {\bf{S}}_i = w_i {\bf{S}}_i.
\end{equation}

Similarity, to update $\mu$ using \eqref{eq_mu1},

\begin{equation}
{\bf{B}}_{i} = w_i {{\bf{B}}}_{i}, {\bf{C}}_{i} = w_i {{\bf{C}}}_{i}.
\end{equation}

Since inliers have much larger weights, $\bf{R}$ and $\mu$ are mainly estimated with inliers.

\subsection{1-Point RANSAC Scheme}

The core algorithm of R1PP$n$P is based on a randomly selected control point $o$. In outlier-free situations, our algorithm works with any control point. However, in situations with outliers, the control point $o$ should be an inlier to make the algorithm work. Hence, we employ the 1-point RANSAC scheme to try different 3D-2D correspondences as the control point until the algorithm finds the correct solution. The 1-point RANSAC scheme combines the core algorithm naturally because the core algorithm can perform the computation with any control point $o \in [1,N]$. We assume that 2D-3D correspondences have the same the possibility to be an inlier, without loss of generality, we select the control point $o$ from the center of all image points to the outside. This is because we found that R1PP$n$P needs less iterations to converge when $o$ is closer to the center, the details of which has been discussed in Section 3.3.

\subsection{Algorithm Flow Chart}
\begin{figure} [ht]
\vspace{0.0cm}
\centering
  \includegraphics[width=.48\textwidth]{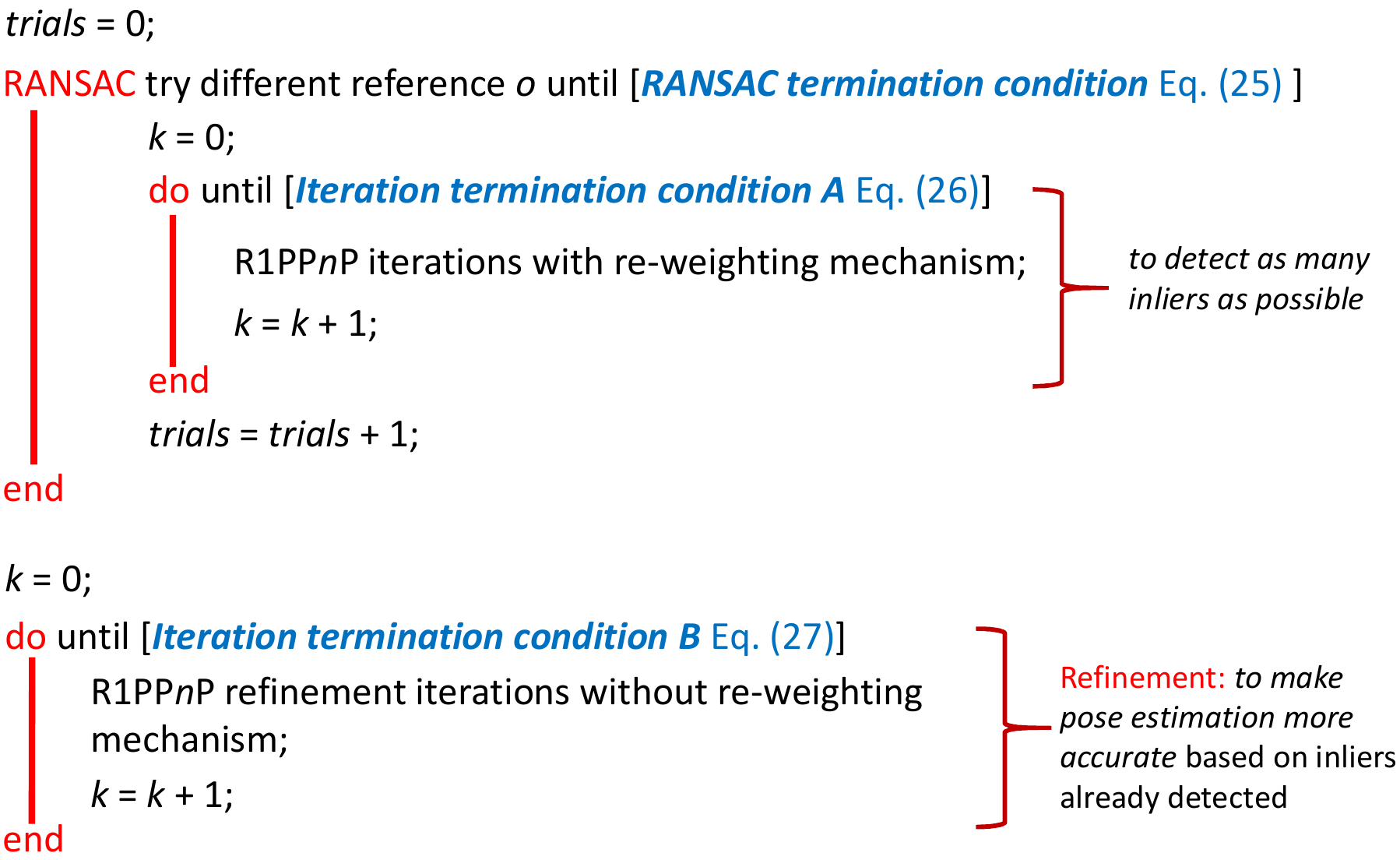}
\caption{The overall flow chart of R1PP$n$P.}
\label{fig_algorithmstep}
\end{figure}

In general, the overall flow chart of R1PP$n$P is shown in Fig. \ref{fig_algorithmstep}, we first detect as many inliers as possible inside the RANSAC framework, then based on the detected inliers, we perform the R1PP$n$P algorithm without re-weighting mechanism to get more accurate results.

Appropriate termination conditions seek balance between speed and precision for RANSAC-based or iterative algorithms. As shown in Fig. \ref{fig_algorithmstep}, two types of termination conditions need to be specified for R1PP$n$P.

\noindent (1) \emph{RANSAC Termination Condition}

The standard RANSAC termination condition \cite{fischler1981random} was employed for R1PP$n$P, that is

\begin{equation}
\label{eq_ransactermination}
trials \ge \log (1 - p)/\log (1 - {p_{\rm{inliers}} ^s}),
\end{equation}

\noindent where $p$ is the certainty and we use $p = 0.99$ for all RANSAC-based methods in this paper, $trials$ is the number of RANSAC trials, $p_{\rm{inliers}} = $ (maximum number of detected inliers) / (number of all points), $s$ is the number of control points needed in each RANSAC trial. $s = 1$ for R1PP$n$P, and $s = 3,4$ for RANSAC+P3P and P4P respectively.

During the RANSAC process, the camera pose estimated by conventional RANSAC+P3P or P4P methods is based on very small number of points. Because of image noise, the estimated pose varies with different inliers as the control points. This is especially serious when the image noise is large. To improve accuracy, the termination condition \eqref{eq_ransactermination} suggests that the standard RANSAC scheme may continue looking for better results after finding a large percentage of inliers. In contrast, R1PP$n$P takes into account all points when estimating the pose, which makes it insensitive to the selected control point $o$. Therefore it is a reasonable assumption that when $p_{\rm{inliers}}$ is large enough (we used the threshold of $60\%$), no improvement can be found and the RANSAC process of R1PP$n$P could be terminated. Accuracy evaluation results in this paper have testified the rationality of this assumption.

\noindent (2) \emph{Termination Conditions for R1PPnP Iterations}

As shown in Fig. \ref{fig_algorithmstep}, we first detect as many inliers as possible and the related \emph{termination condition A} is satisfied when the detected number of inliers becomes stable, that is,

\begin{equation}
\label{eq_terminationA}
N_{\rm{inlier}}^{(k)} - N_{\rm{inlier}}^{(k-20)} \le 0 \text{  and  } k>20,
\end{equation}

\noindent where $k$ is the index of iterations, $N_{\rm{inlier}}$ is the number of detected inliers. According to our experience, in most cases no more inliers would be detected if $N_{\rm{inlier}}$ has not increased in 20 iterations.

\begin{figure} [htp]
\vspace{0.0cm}
\centering
  \includegraphics[width=.24\textwidth]{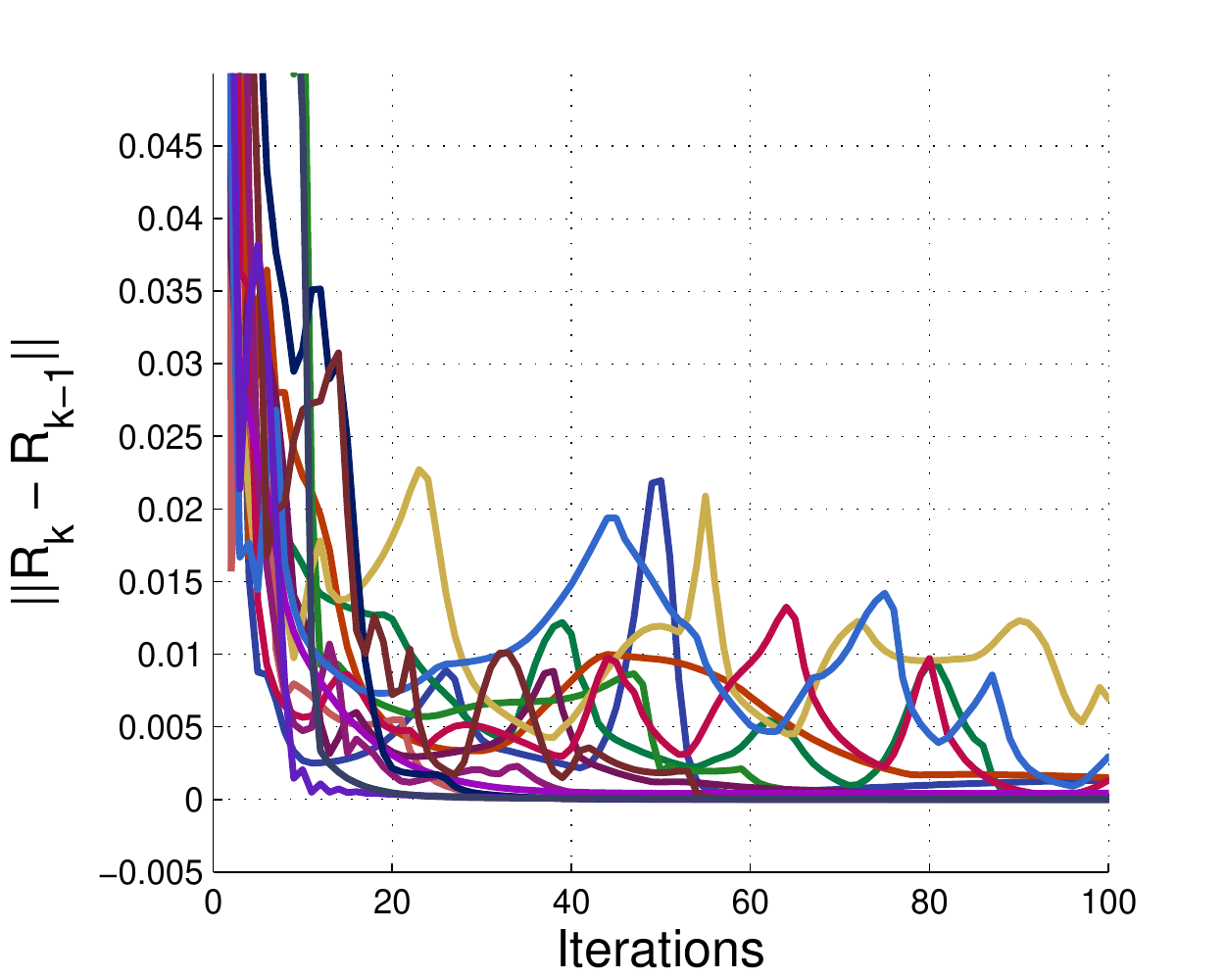}
  \includegraphics[width=.24\textwidth]{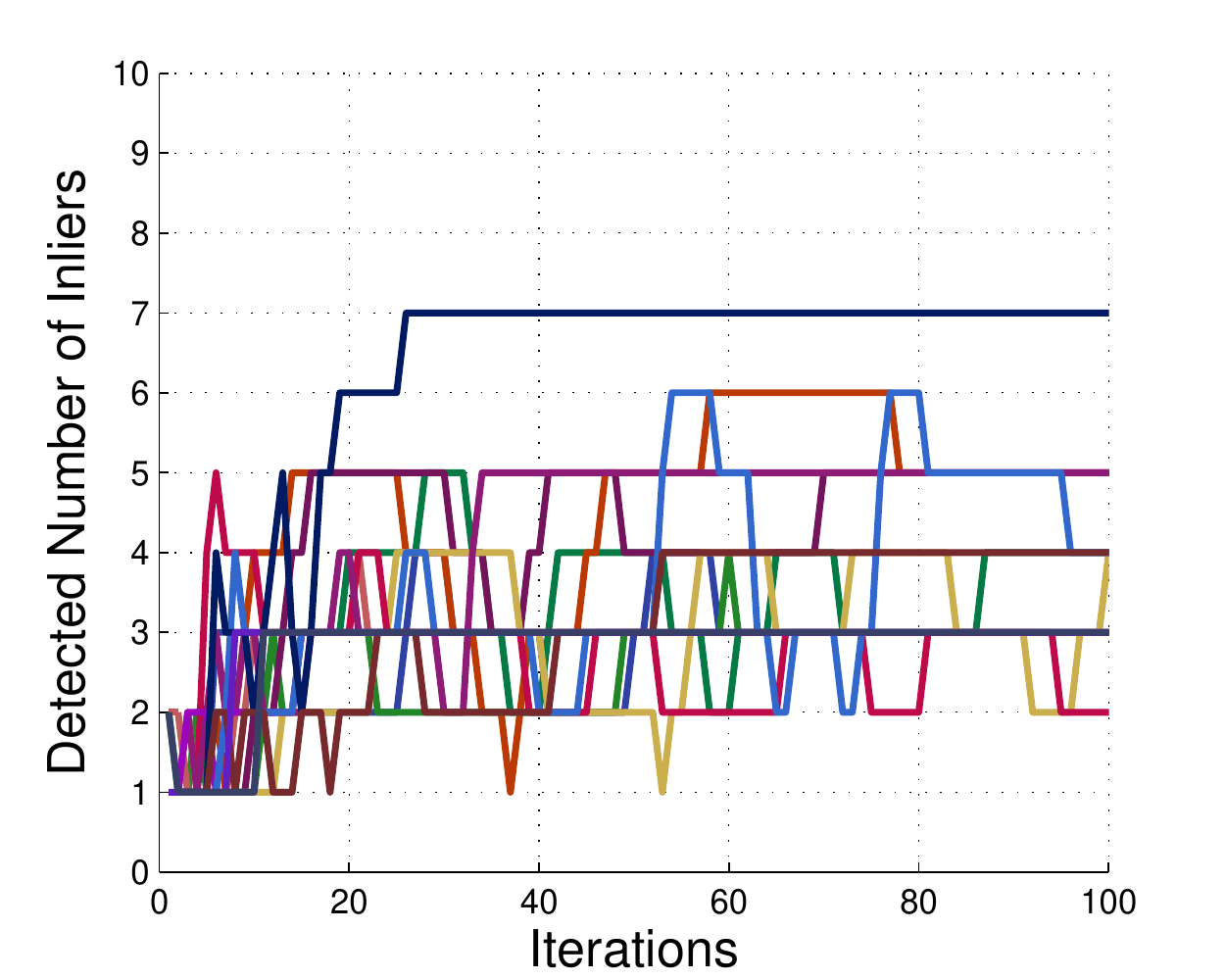}
\caption{Experiments with synthetic data (ordinary 3D case, $50\%$ of outliers) to demonstrate the iteration process of R1PP$n$P when the control point $o$ is an outlier. Randomly colored lines are results with different control points. (a) The changes of estimated camera pose between frames $k$ and $k-1$ are complex during the iteration process, based on which it is difficult to decide when to stop the process. (b) It is more robust and efficient to stop the process when no more inliers can be detected.}
\label{fig_testTerminationA}
\end{figure}

A good \emph{termination condition A} should be able to stop the iteration process as early as possible when point $o$ is an outlier, and do not interrupt when point $o$ is an inlier. We proposed the \emph{termination condition A} with a window size = 20 iterations to seek balance between speed and robustness. This termination condition is not based on the comparison of parameters of adjacent iterations because in R1PP$n$P, the dynamically updated weights $w_i$ may make the convergence process complex, especially when point $o$ is an outlier. As shown in Fig. \ref{fig_testTerminationA}(a), with an outlier as the control point, $\left\| {{{\bf{R}}^{(k)}} - {{\bf{R}}^{(k - 1)}}} \right\|$ may take many iterations to converge to zero, which is slow. With the change of detected number of inliers in a larger window size, the termination decision can be more robust and efficient, as shown in Fig. \ref{fig_testTerminationA}(b).

The refinement stage makes pose estimation results more accurate based on the detected inliers. Without the reweighting mechanism, the convergence process is much simple. Hence the \emph{termination condition B} is satisfied when the estimated rotation becomes stable, that is,

\begin{equation}
\label{eq_terminationB}
\left\| {{{\bf{R}}^{(k)}} - {{\bf{R}}^{(k - 1)}}} \right\| < 1e-5.
\end{equation}

\section{Experiments}

\begin{figure} [htp]
\vspace{0.0cm}
\centering
  \includegraphics[width=.24\textwidth]{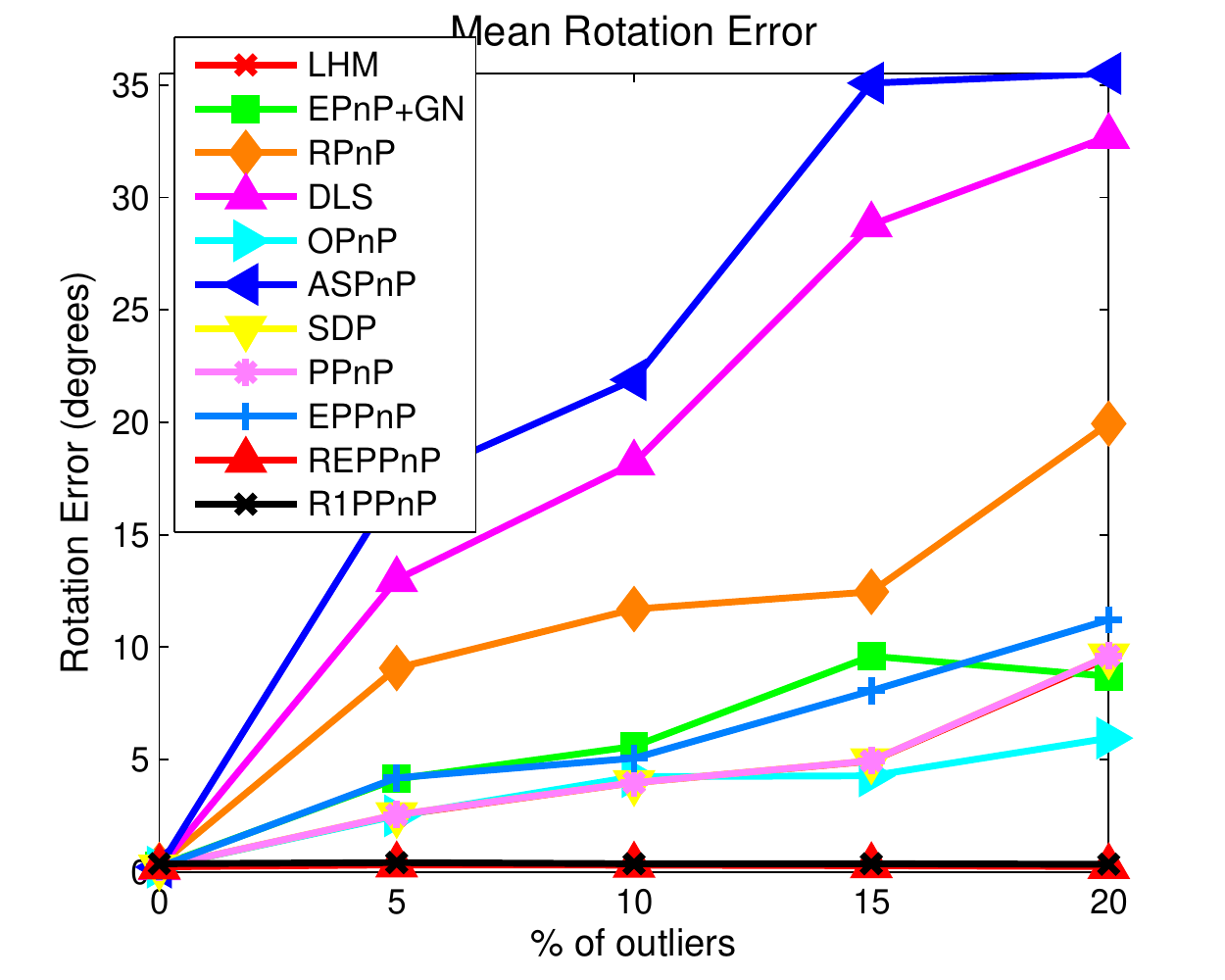}
  \includegraphics[width=.24\textwidth]{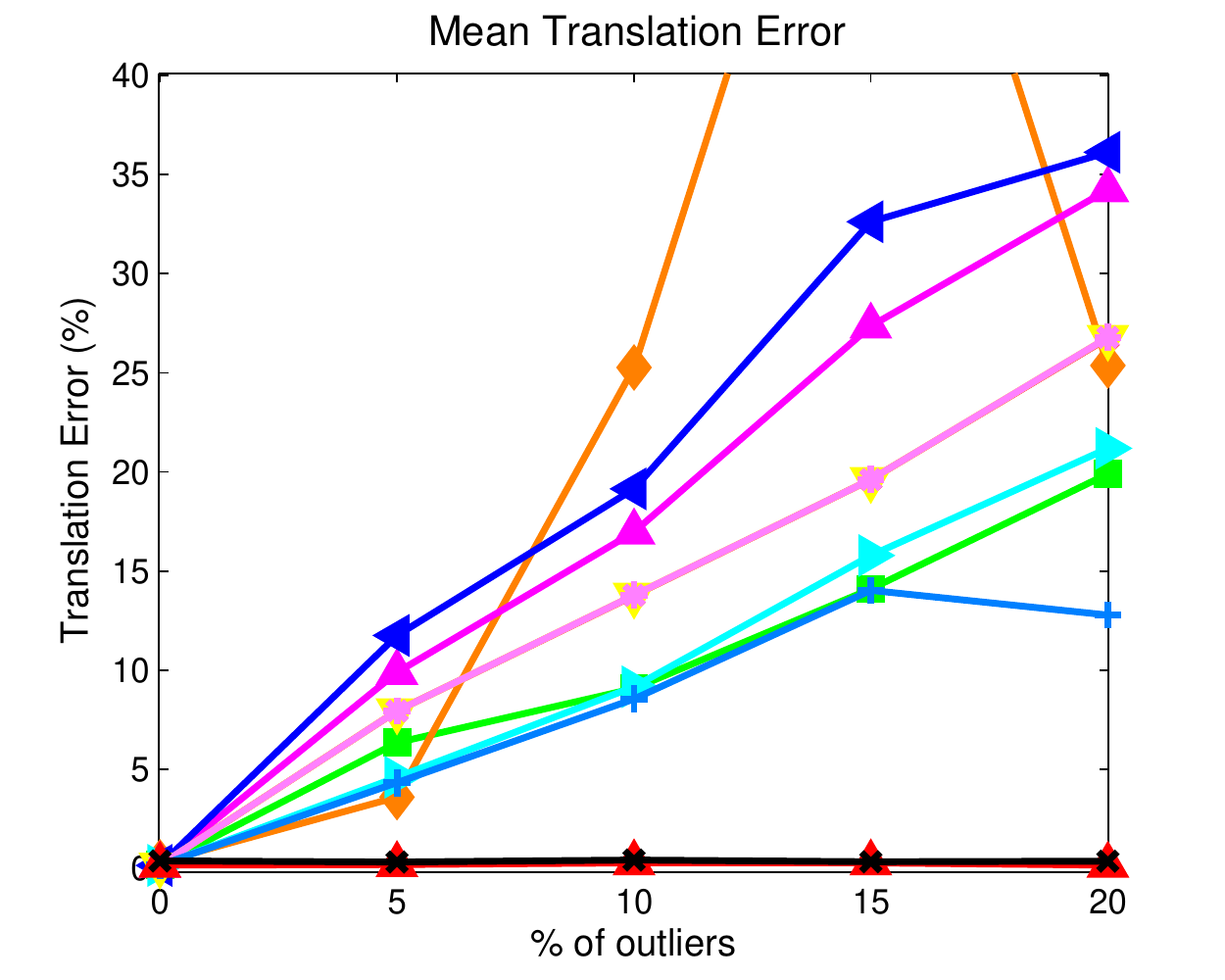}
\caption{Except for REPP$n$P and R1PP$n$P, most P$n$P methods cannot handle outliers.}
\label{fig_pnpcannothandleoutliers}
\end{figure}

\begin{figure*} [htp]
\vspace{0.0cm}
\centering
  \includegraphics[width=.42\textwidth]{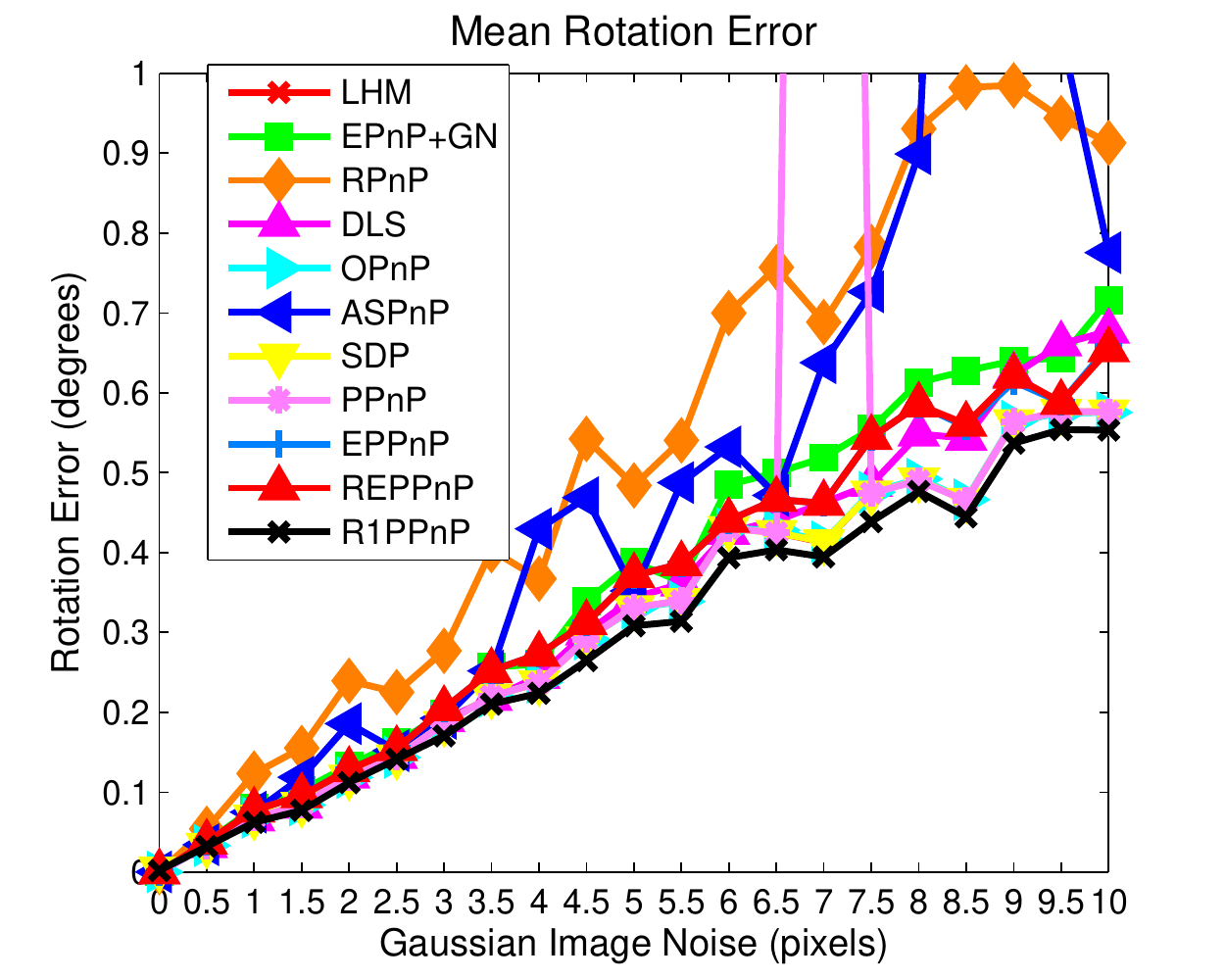}
  \includegraphics[width=.42\textwidth]{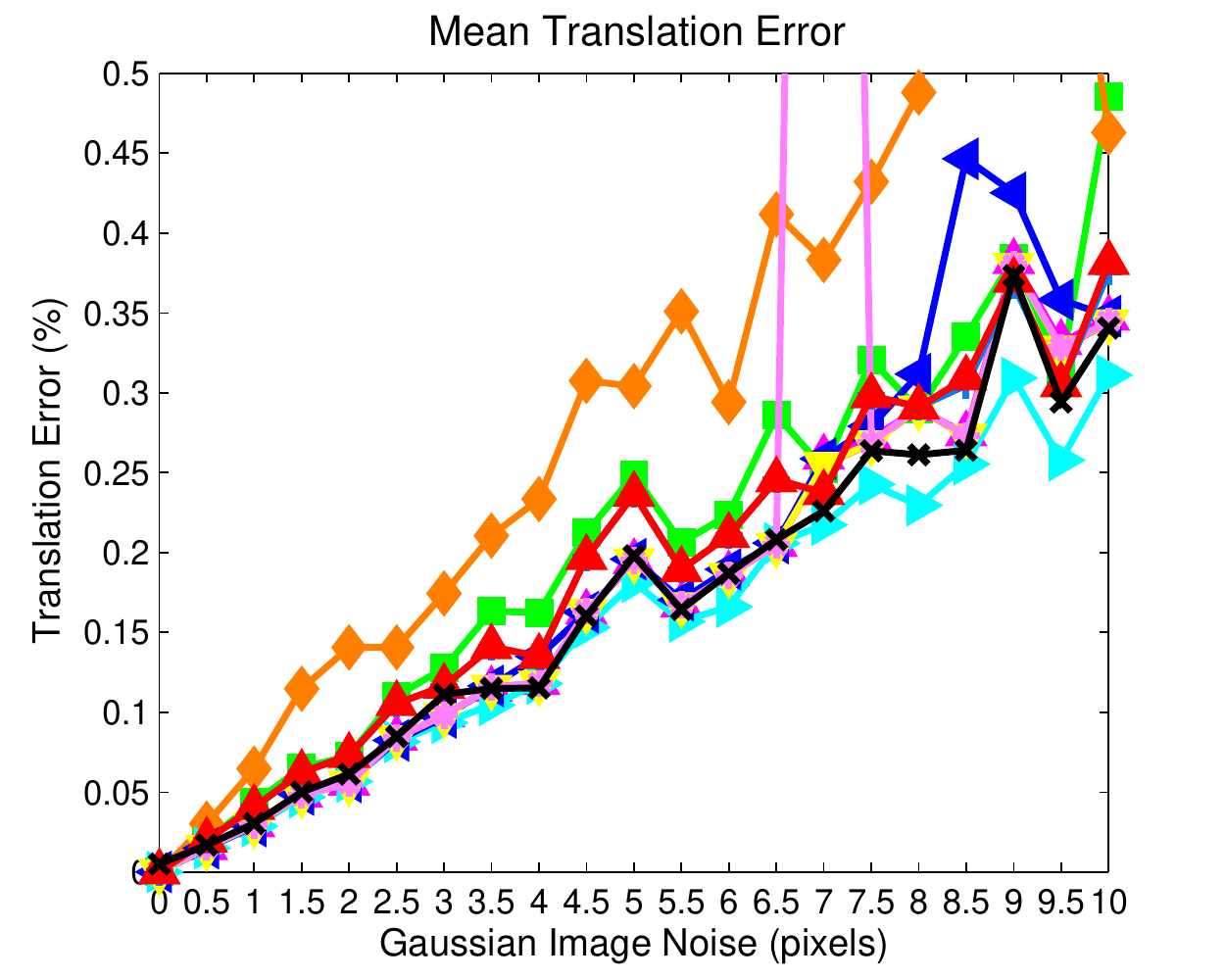}
\caption{Accuracy with outlier-free synthetic data (ordinary 3D cases). Number of points was 100. Different levels of image noises were added.}
\label{fig_nooutlers_accuracy_ordinary}
\end{figure*}

\begin{figure*} [htp]
\vspace{0.0cm}
\centering
  \includegraphics[width=.42\textwidth]{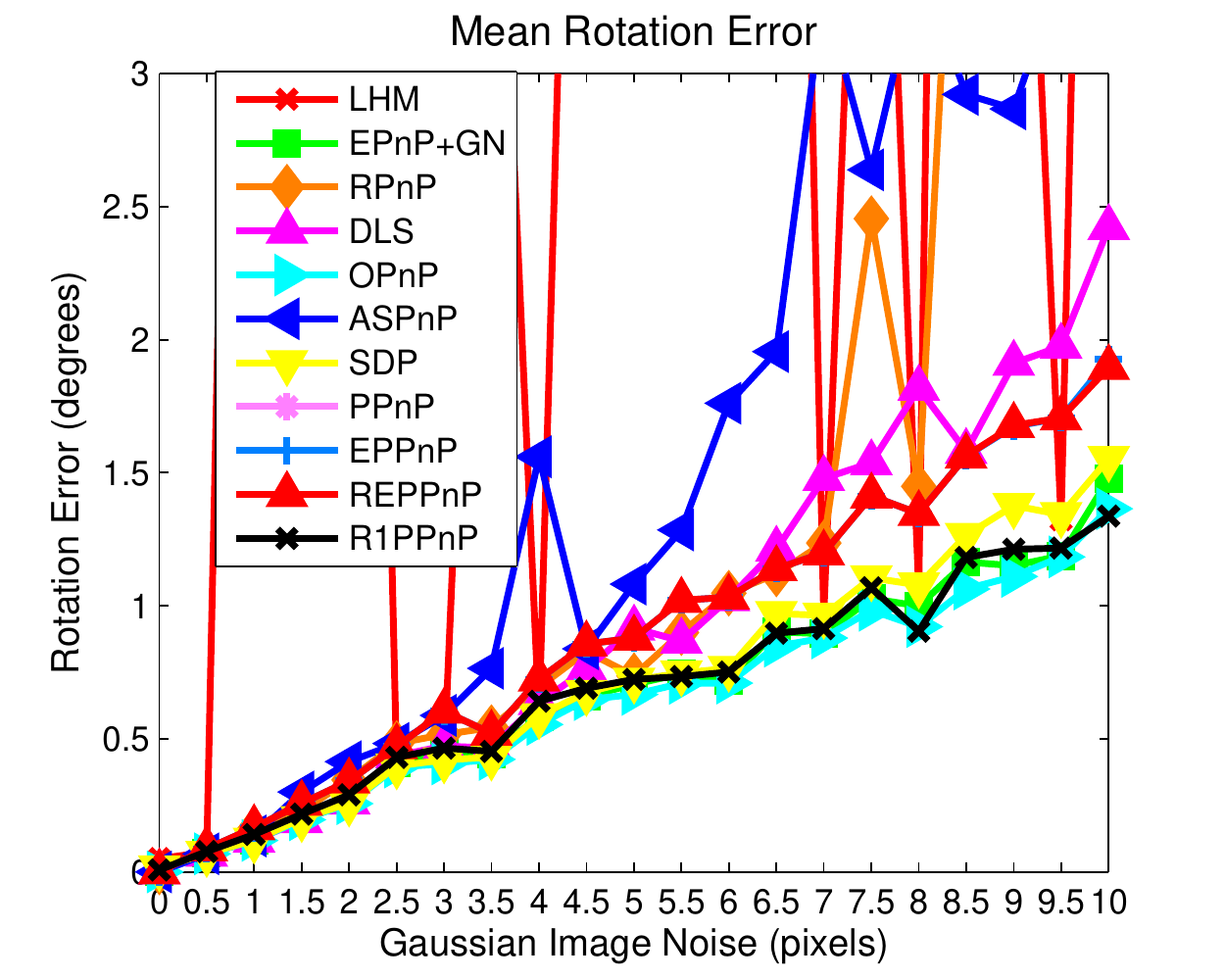}
  \includegraphics[width=.42\textwidth]{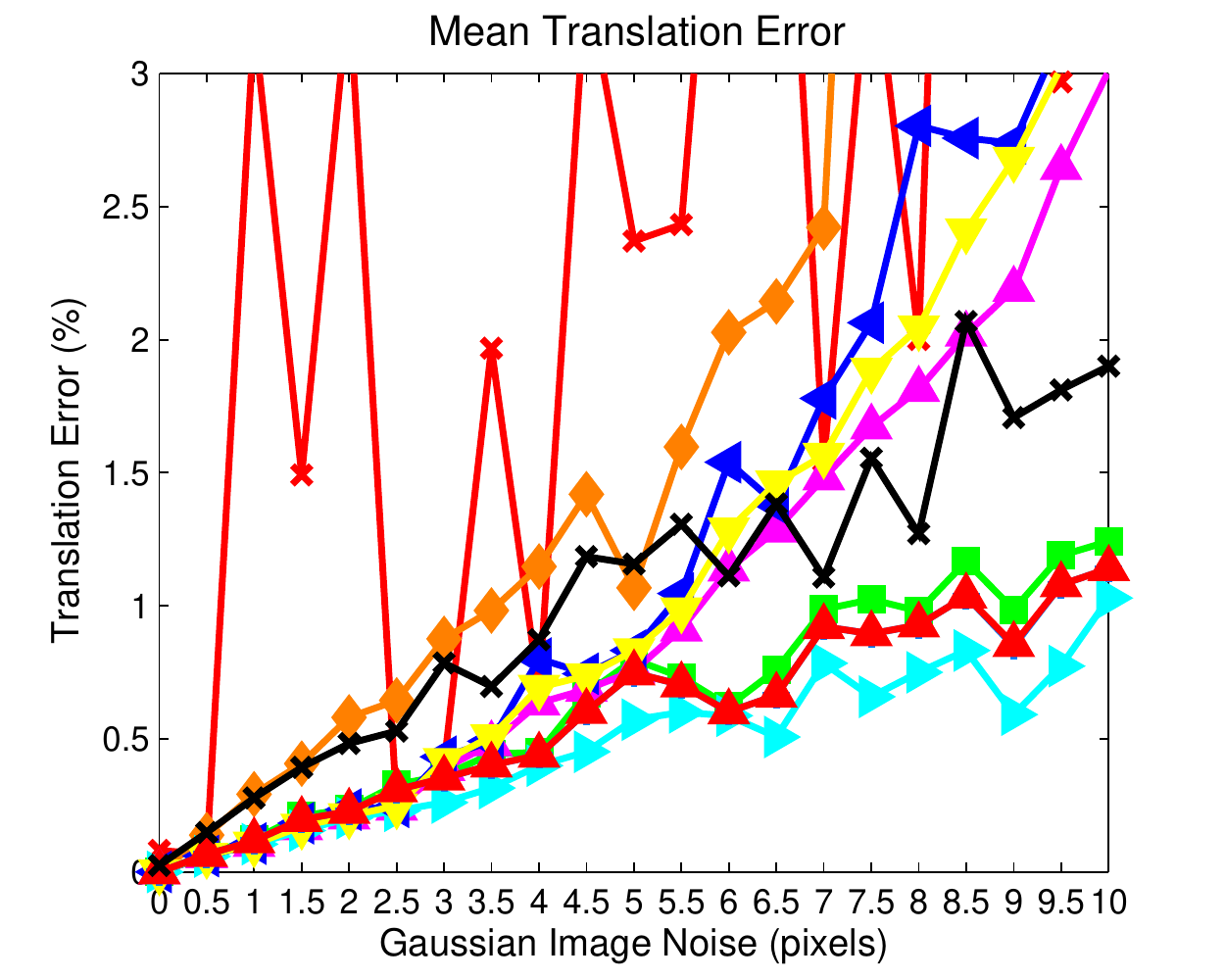}
\caption{Accuracy with outlier-free synthetic data (quasi-singular cases). Number of points was 100. Different levels of image noises were added. PPnP is out of range. The accuracy of all P$n$P methods decreased significantly compared with those in ordinary 3D cases, as shown in Fig.\ref{fig_nooutlers_accuracy_ordinary}. }
\label{fig_nooutlers_accuracy_quasi}
\end{figure*}

\begin{figure*} [htp]
\vspace{0.0cm}
\centering
  \includegraphics[width=.42\textwidth]{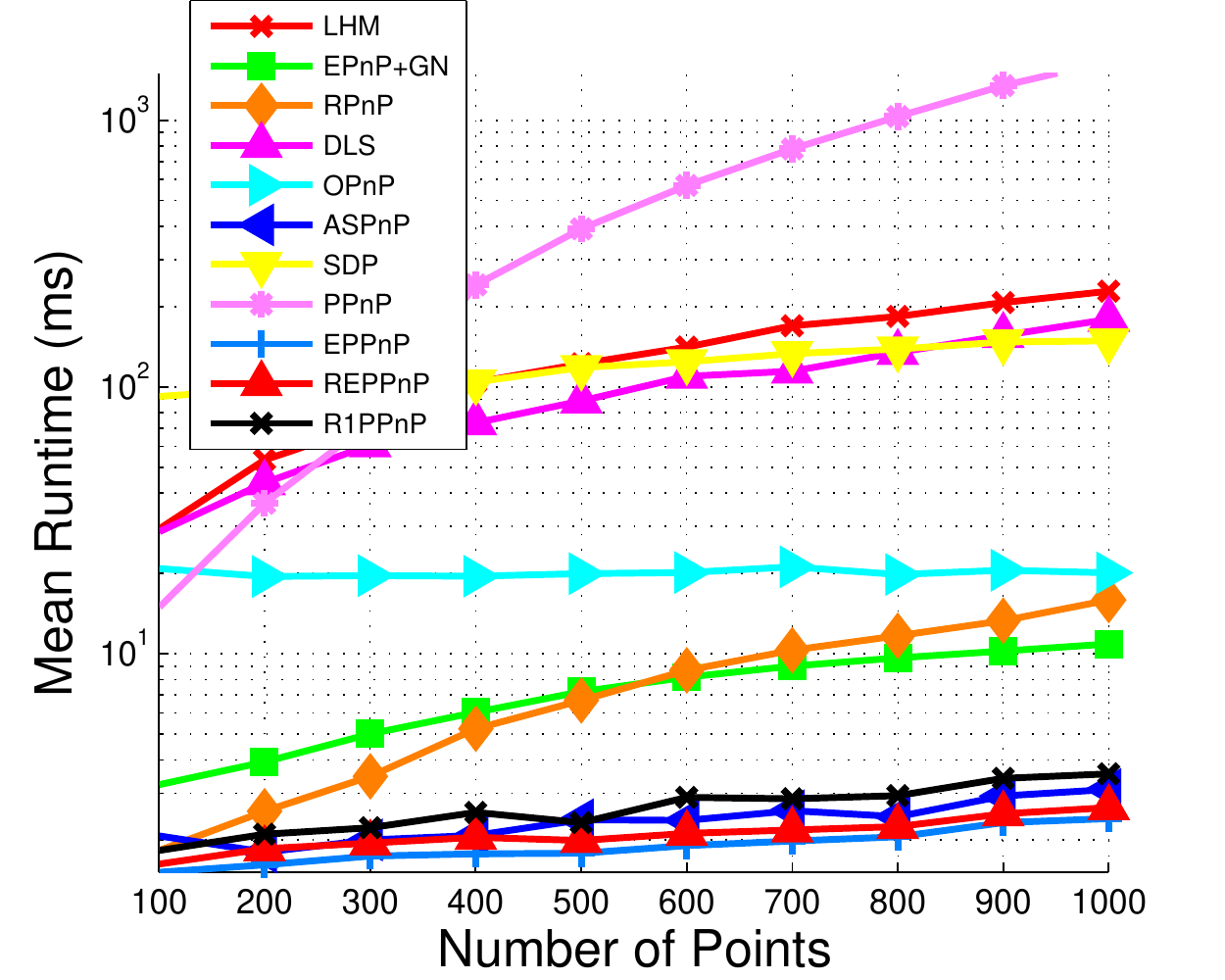}
  \includegraphics[width=.42\textwidth]{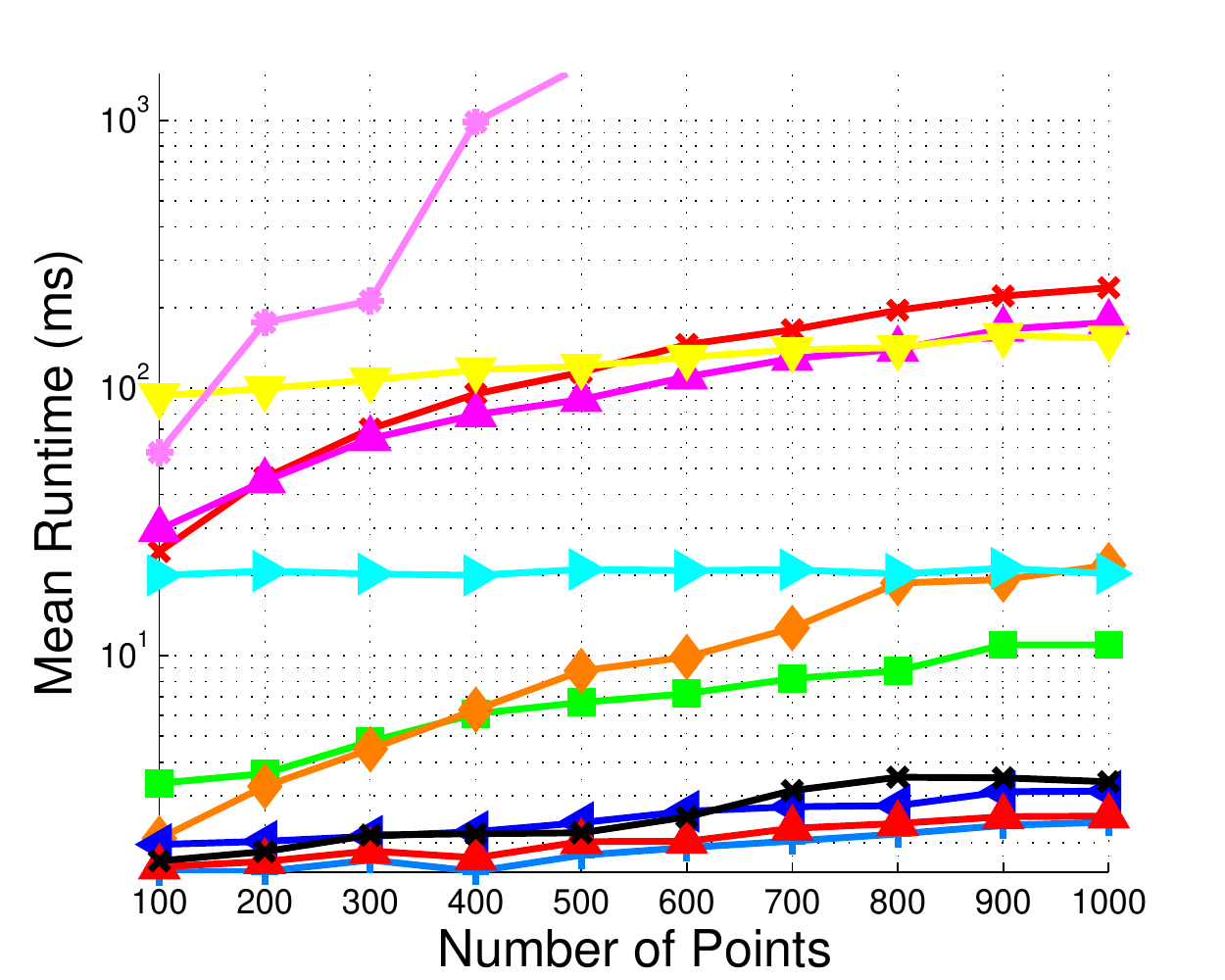}
\caption{Runtime results with outlier-free synthetic data. Standard deviation of image noise $\sigma=5$ pixels. The number of points increased from 100 to 1000. (Left) Ordinary 3D cases. (Right) Quasi-singular cases.}
\label{fig_nooutlers_runtime}
\end{figure*}

The performance of the proposed R1PP$n$P algorithm was evaluated by comparing against the state-of-the-art P$n$P methods. The source code was implemented in MATLAB scripts and executed on a computer with an Intel Core i7 2.60 GHz CPU. We used both synthetic and real-world data to conduct evaluation experiments. The initial values for R1PP$n$P are ${\bf{R}} = diag\left \{ 1,1,1\right \} $ and $\mu = 1e-4$. RANSAC+P3P or P4P methods also used the standard termination condition \eqref{eq_ransactermination}.


\subsection{Synthetic Experiments}

 Synthetic experiments in this paper shared the following parameters. The camera focal length is 1,000 pixels with a resolution of $640\times480$. Two types of synthetic data were generated. (1) \textbf{Ordinary three-dimensional (3D) case}: object points were randomly and uniformly distributed in a cube region $[-2,2]\times[-2,2]\times[4,8]$. (2) \textbf{Quasi-singular case}: The distribution cube is $[1,2]\times[1,2]\times[4,8]$. For each experiment result, we report the mean values of 100 trials.

For accuracy evaluation, the rotation error is measured in degrees between the truth rotation $\bf{R}_{\rm{true}}$ and the estimated $\bf{R}$ as ${e_{{\rm{rot}}}}(\deg ) = \left\| {{{\left[ {{\rm{acos}}{{({\bf{r}}_{k,{\rm{true}}}^T \cdot {{\bf{r}}_k})}_{k = 1,2,3}}} \right]}^T}} \right\| \times {\rm{180/}}\pi  $, where ${{\bf{r}}_{k,{\rm{true}}}}$ and ${\bf{r}}_k$ are the $k$th column of $\bf{R}_{\rm{true}}$ and $\bf{R}$ respectively. The translation error is ${e_{{\rm{trans}}}}(\% ) = \left\| {{{\bf{t}}_{{\rm{true}}}} - {\bf{t}}} \right\|/\left\| {\bf{t}} \right\| \times 100\%$.

\subsubsection{Outlier-Free Synthetic Situations}

Most P$n$P algorithms do not have the ability to handle outliers, and even a small percentage of outliers will significantly reduce the accuracy, as shown in Fig.\ref{fig_pnpcannothandleoutliers}. Thus, although outlier-free situations are not the main concern of R1PP$n$P, we first conducted comparison experiments between the proposed R1PP$n$P and other P$n$P algorithms in outlier-free situations. The reason for this comparison is that RANSAC+P3P or P4P methods usually need other P$n$P methods as the final refinement step. Hence the accuracy and speed in outlier-free situations are also related to the performance in situations with outliers.

Here we only performed the core algorithm of R1PP$n$P without outliers handling mechanism. The termination condition for R1PP$n$P iterations was as Eq.\eqref{eq_terminationB}. In this experiment, we compared our proposed R1PP$n$P with the following P$n$P methods: LHM \cite{lu2000fast}, EP$n$P \cite{lepetit2009epnp}, RP$n$P \cite{li2012robust}, DLS \cite{hesch2011direct}, OP$n$P \cite{zheng2013revisiting}, ASP$n$P \cite{zheng2013aspnp}, SDP \cite{schweighofer2008globally}, PP$n$P \cite{garro2012solving}, EPP$n$P \cite{ferraz2014very} and REPP$n$P \cite{ferraz2014very}.

\begin{figure*} [htp]
\vspace{0.0cm}
\centering
 \subfigure[]{
  \includegraphics[width=.42\textwidth]{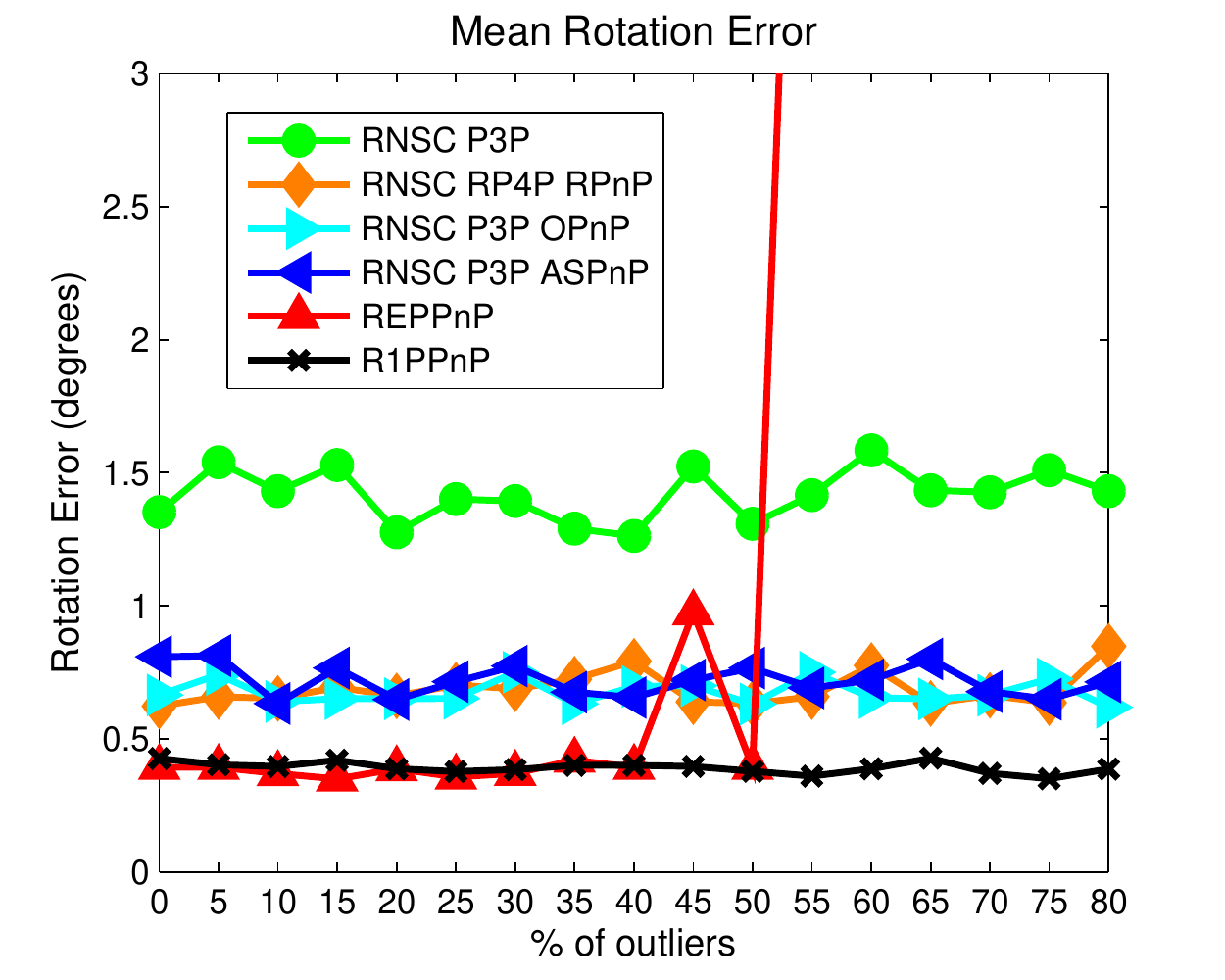}
    }
 \subfigure[]{
  \includegraphics[width=.42\textwidth]{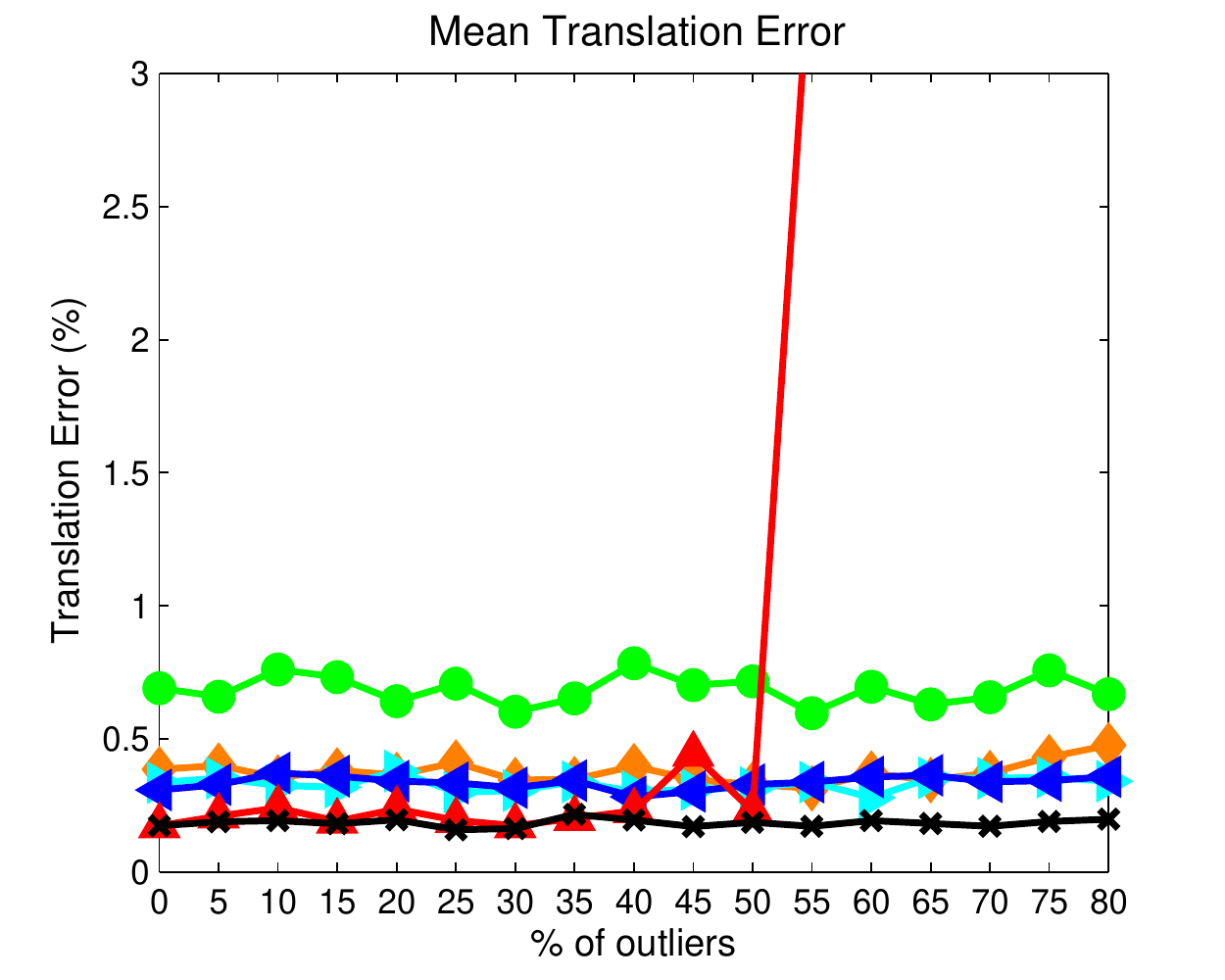}
    }
 \subfigure[]{
  \includegraphics[width=.42\textwidth]{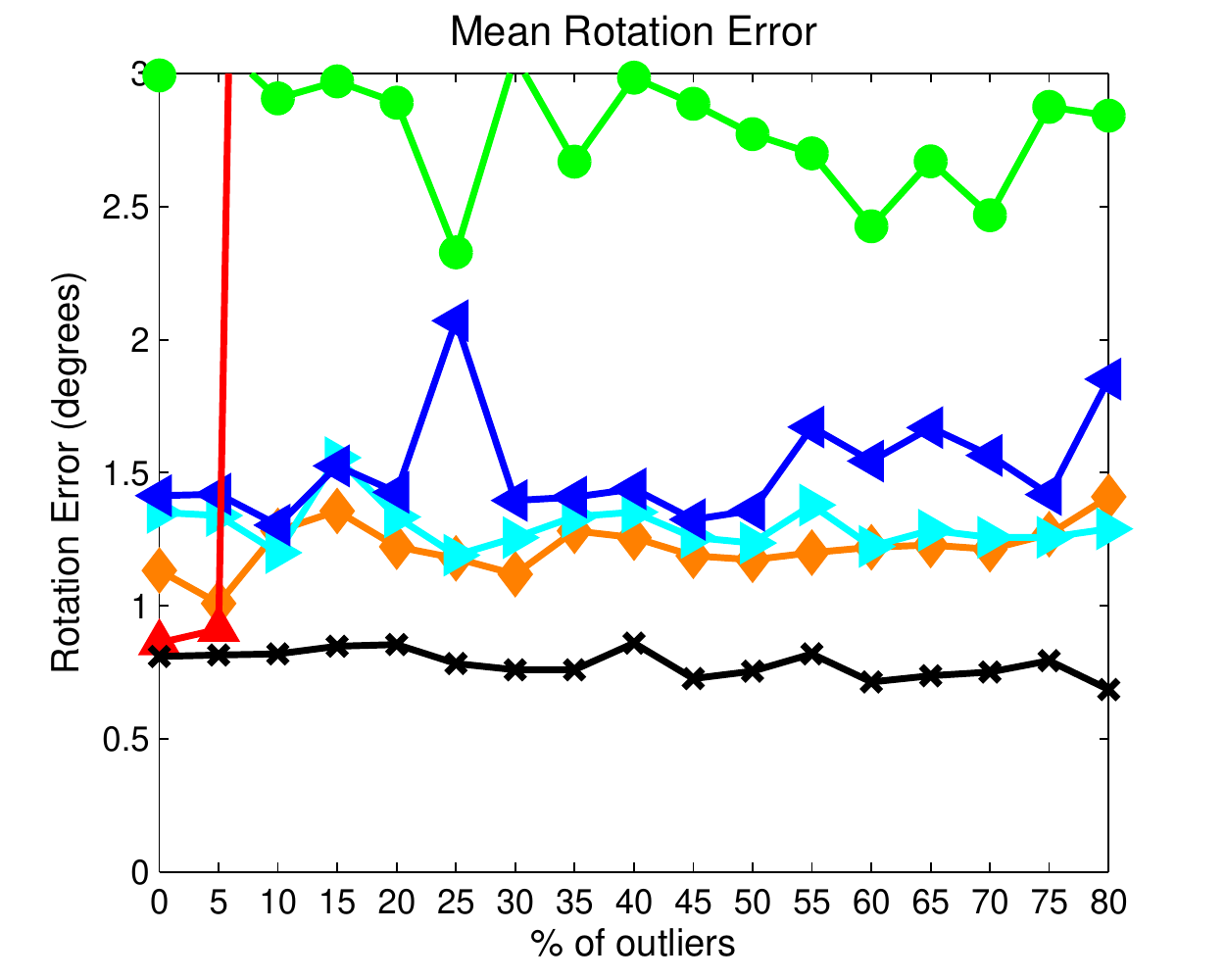}
    }
 \subfigure[]{
  \includegraphics[width=.42\textwidth]{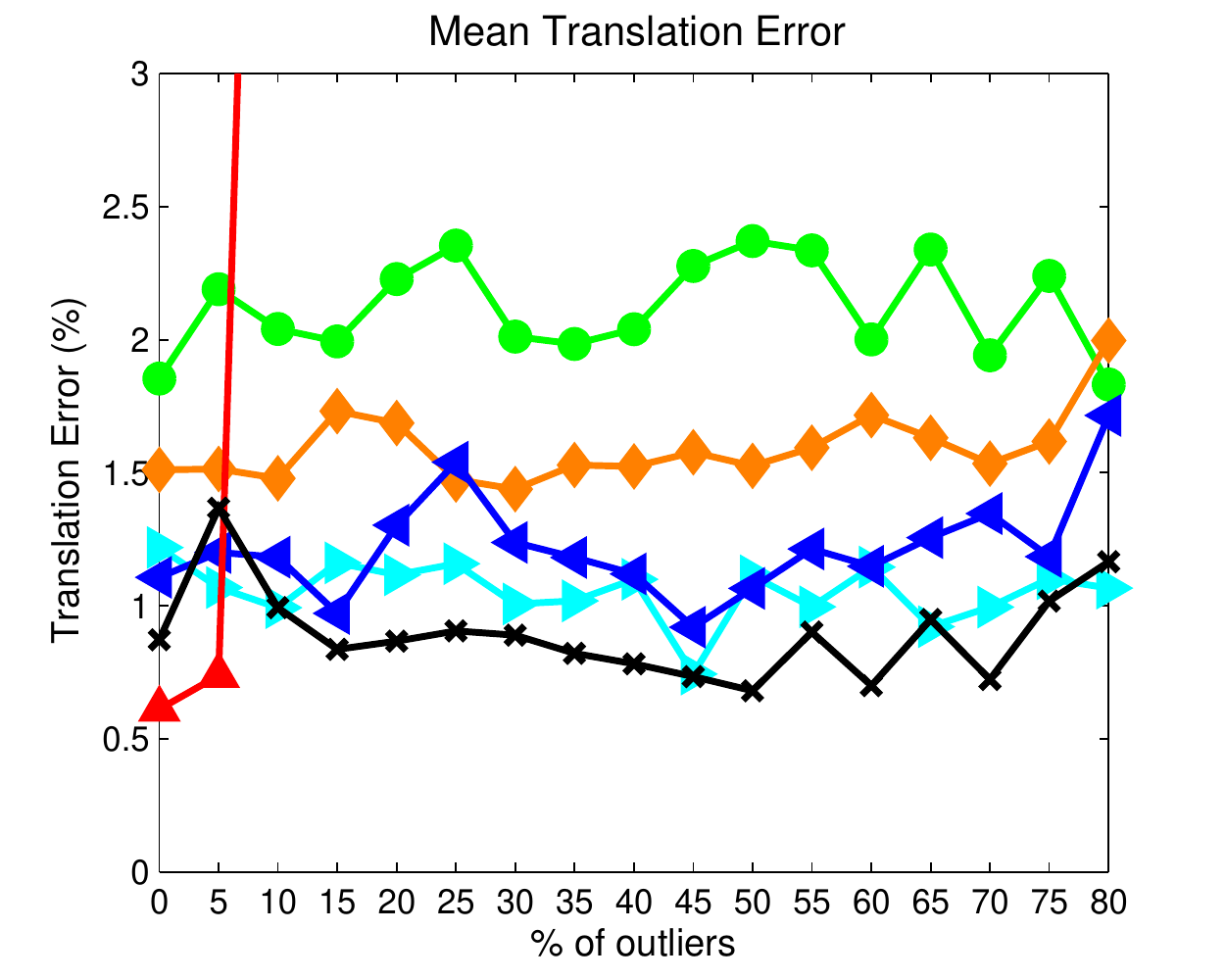}
    }
  \caption{Average accuracy on synthetic data with outliers. (a)-(b) Accuracy with ordinary 3D cases; (c)-(d) Accuracy with quasi-singular cases.}
\label{fig_outliers}
\end{figure*}

\begin{figure*} [!htb]
\vspace{0.0cm}
\centering
  \includegraphics[width=.42\textwidth]{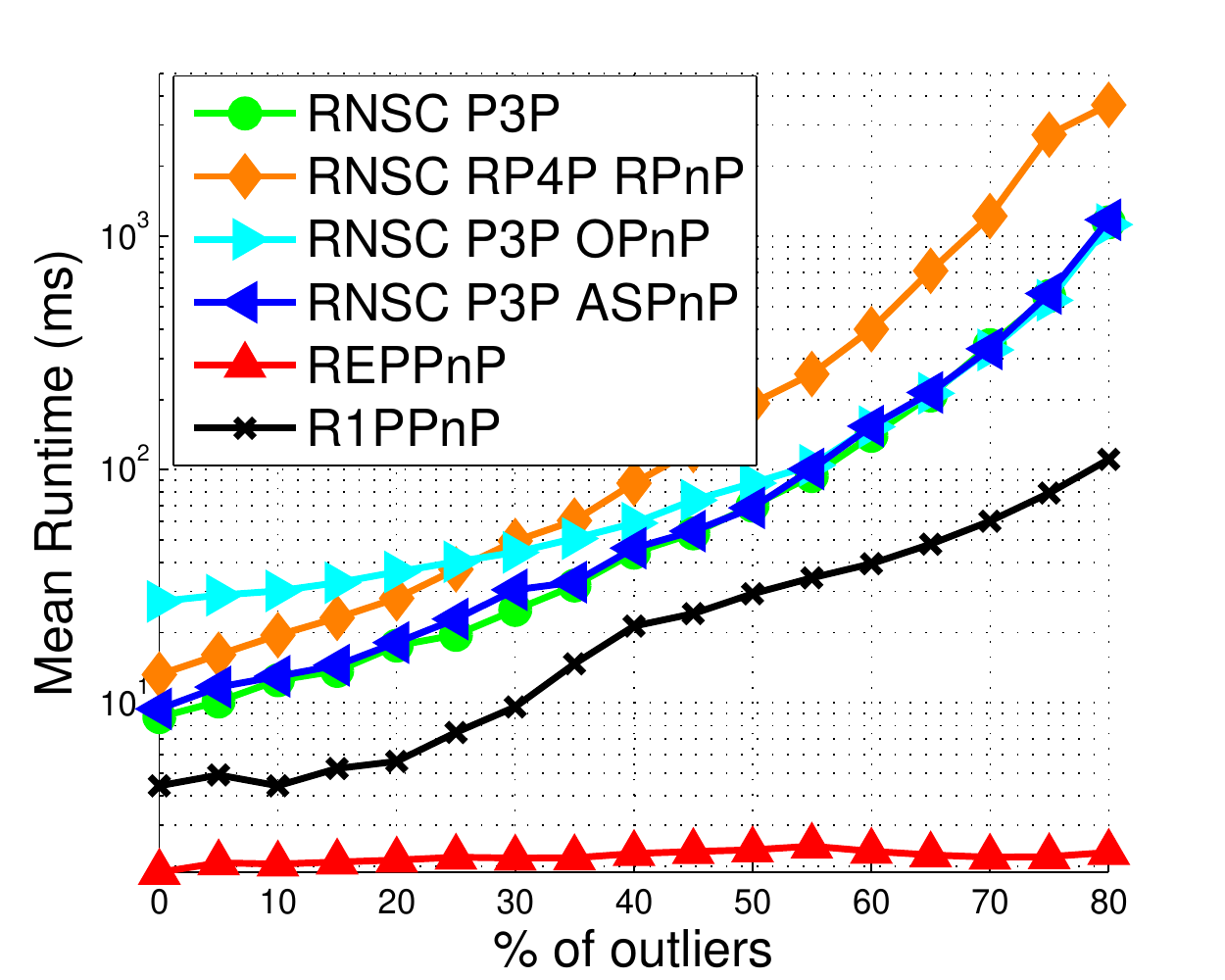}
  \includegraphics[width=.42\textwidth]{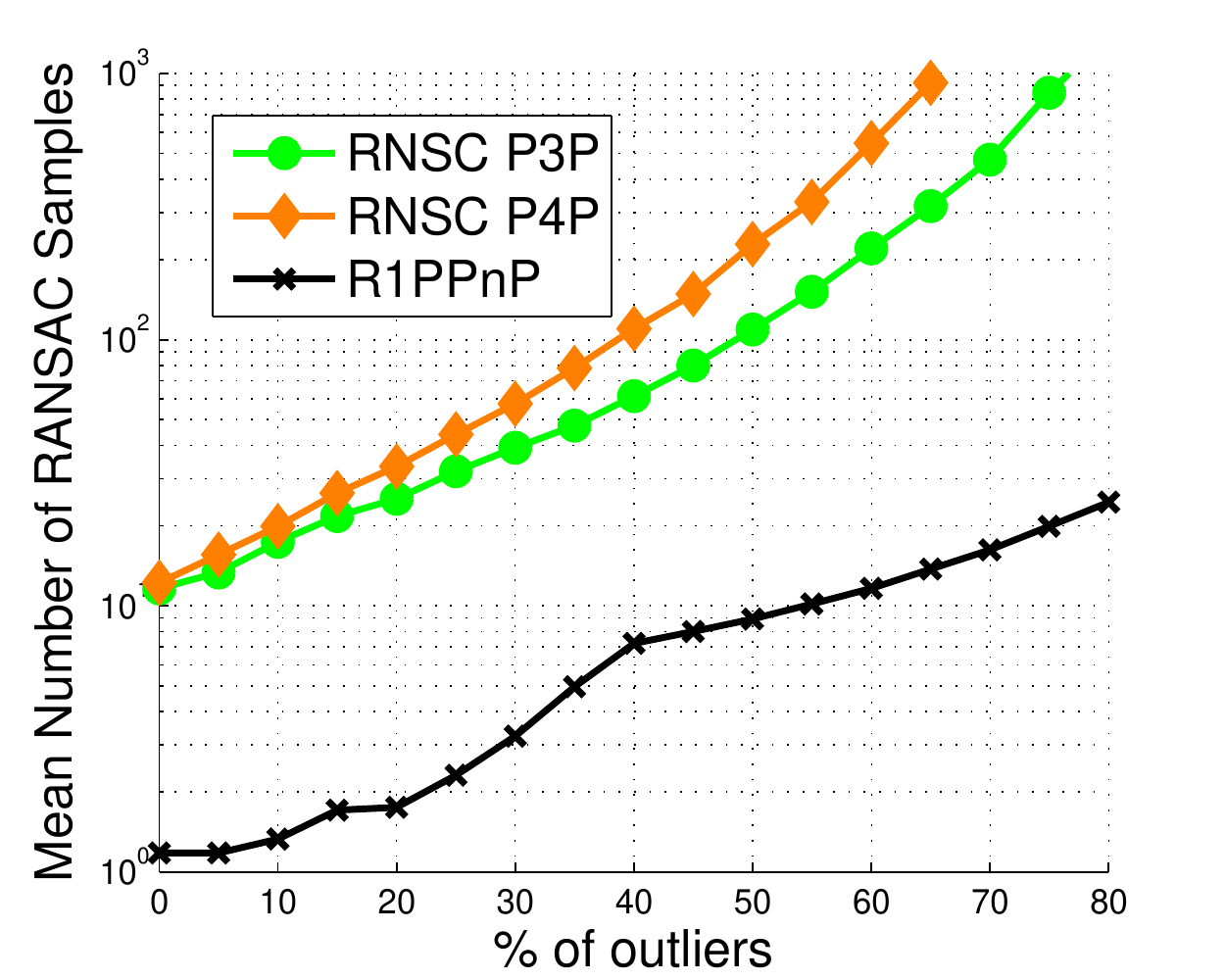}

  \caption{Average runtime and number of required RANSAC samples with ordinary 3D cases synthetic data.  We do not give the results with quasi-singular cases because they is very close to that with ordinary 3D cases. RANSAC+P3P or P4P needs more than 10 RANSAC trails when $p_{\rm{outliers}}=0$ because the large image noise ($\sigma = 5$ pixels) usually makes P3P or P4P methods unable to find the correct pose with 3 or 4 inliers to satisify the termination condition Eq.\eqref{eq_ransactermination}. In contrast, the required number of RANSAC trials of R1PP$n$P is not sensitive to image noise because all points are taken into account.}
\label{fig_outlierstime}
\end{figure*}

In our accuracy evaluation experiments, the number of points was 100 and we added different levels of Gaussian image noises from 0 to 10 pixels. As shown in Figs. \ref{fig_nooutlers_accuracy_ordinary} and \ref{fig_nooutlers_accuracy_quasi}, for both ordinary 3D and quasi-singular cases, R1PP$n$P gave the most accurate rotation estimation results together with OP$n$P and SDP. For ordinary 3D cases, R1PP$n$P was among the most accurate methods to estimate translation and was only sightly less accurate than OP$n$P. However, for quasi-singular cases, the accuracy of translation estimation of R1PP$n$P was not the state-of-the-art. ASP$n$P became unstable with large image noise hence its mean accuracy decreased significantly compared with that with small image noise. Although sometimes PP$n$P can provide accurate rotation estimation results in ordinary 3D cases, it also suffered from  instability in some random cases, as shown by the jitter in Fig. \ref{fig_nooutlers_accuracy_ordinary}. PP$n$P and LHM cannot handle the quasi-singular cases.

To evaluate runtime, Gaussian image noise with a standard deviation of $\sigma=5$ pixels was added and the number of points increased from 100 to 1000. As shown in Fig. \ref{fig_nooutlers_runtime}, the proposed R1PPnP, together with EPP$n$P, REPP$n$P and ASP$n$P showed superior computational speed. The runtime of R1PP$n$P did not grow significantly with respect to the number of points. We suspect that this results from the intrinsic parallel optimization of the matrix computation of MATLAB 2014a.

Generally speaking, in outlier-free situations, R1PP$n$P was among the state-of-the-art methods in terms of both accuracy and computational speed. One drawback of R1PP$n$P is that the accuracy of translation estimation in quasi-singular cases was not among the best.

\subsubsection{Synthetic Situations with Outliers}

\begin{figure*} [htp]
\vspace{0.0cm}
\centering

  \subfigure[]{
  \includegraphics[width=.45\textwidth]{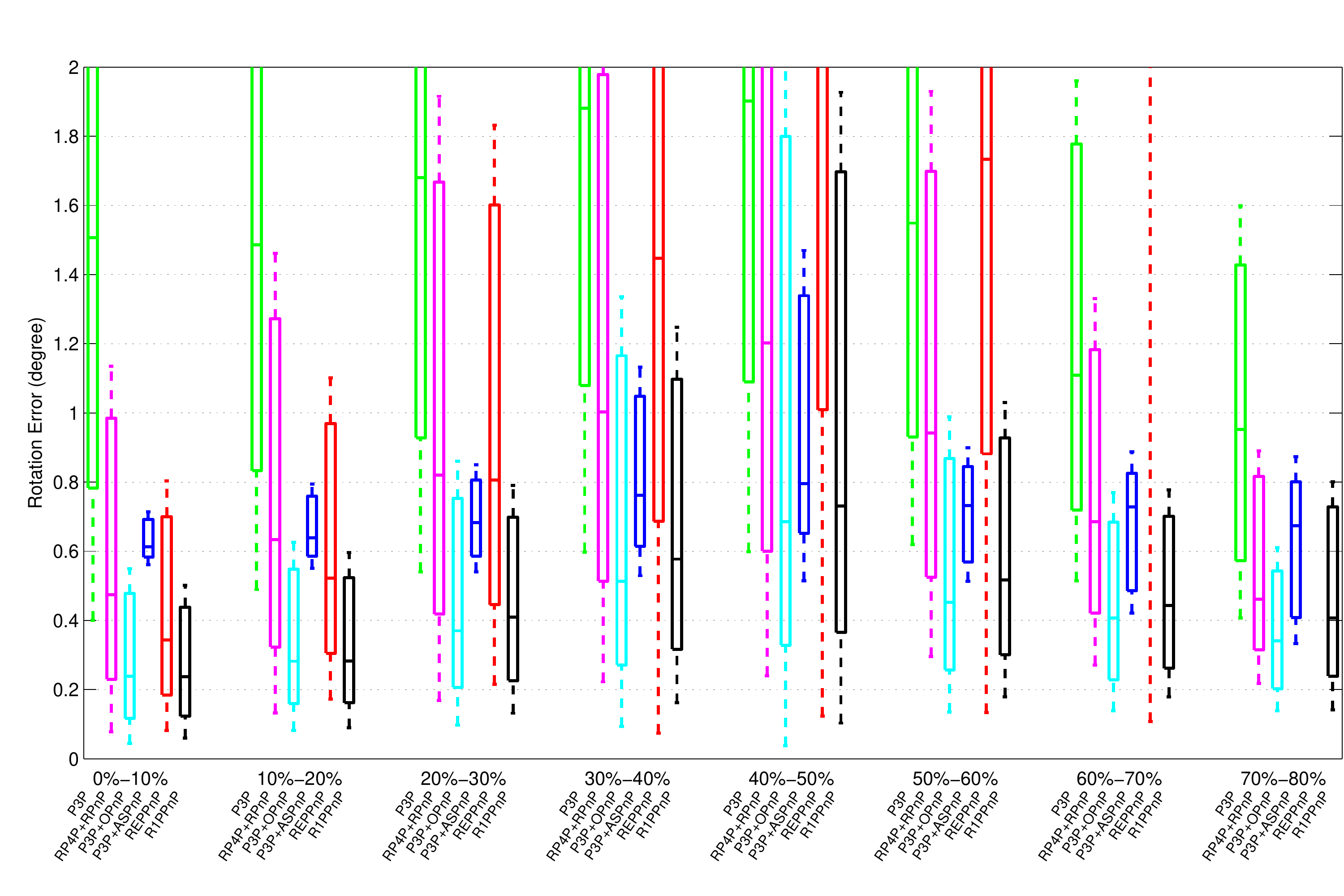}
  }
  \subfigure[]{
  \includegraphics[width=.45\textwidth]{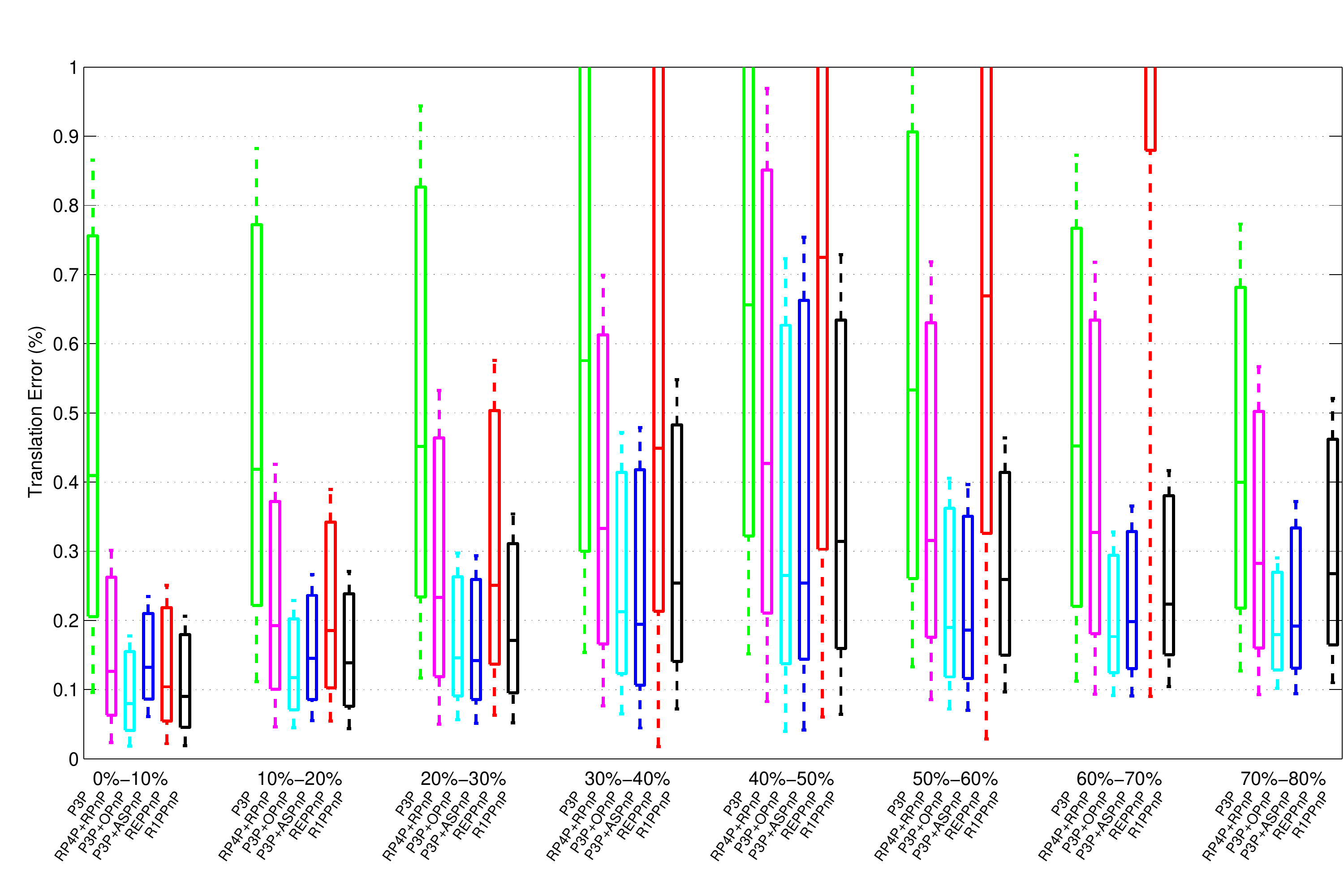}
  }
  \subfigure[]{
  \includegraphics[width=.45\textwidth]{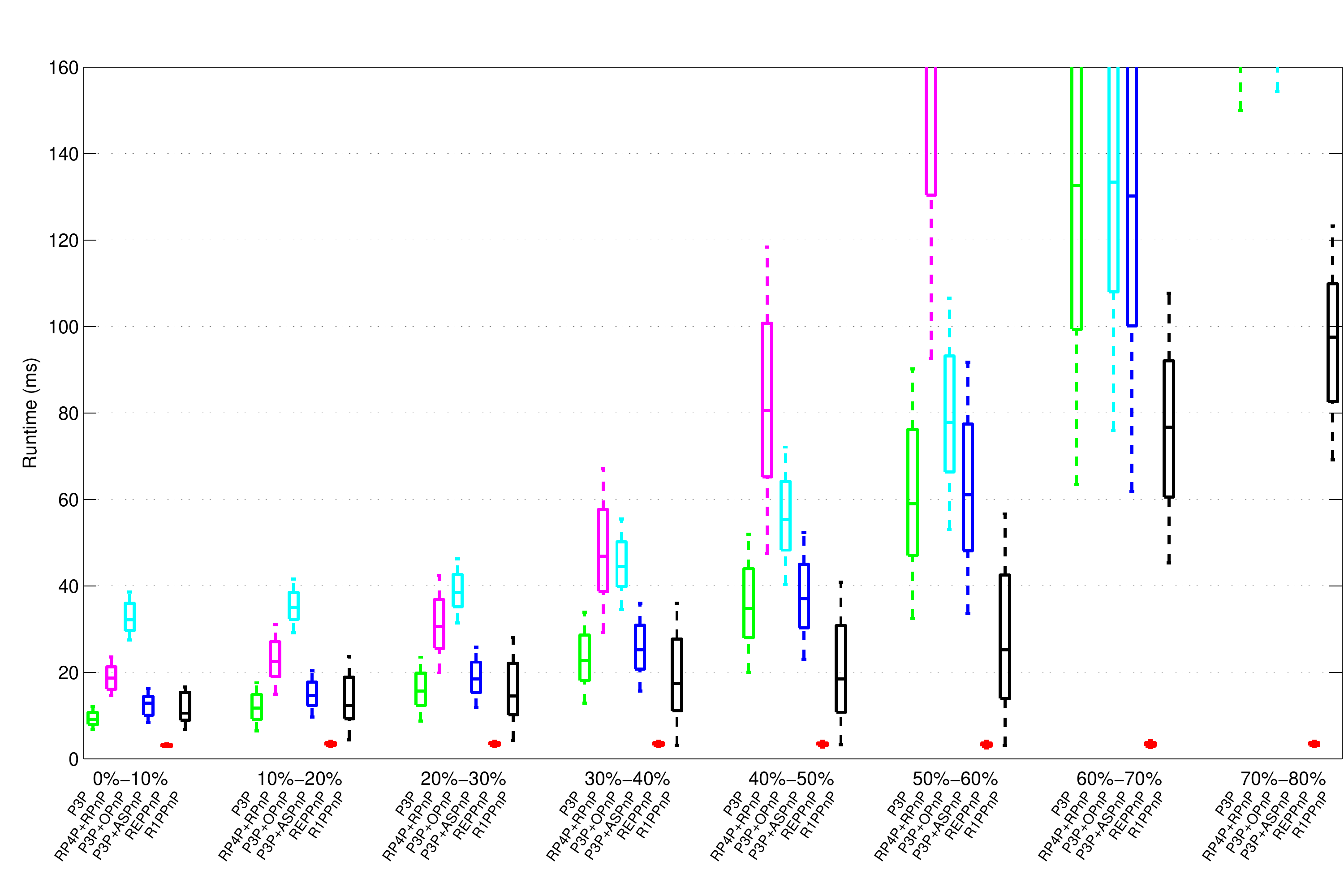}
  }
  \subfigure[]{
  \includegraphics[width=.45\textwidth]{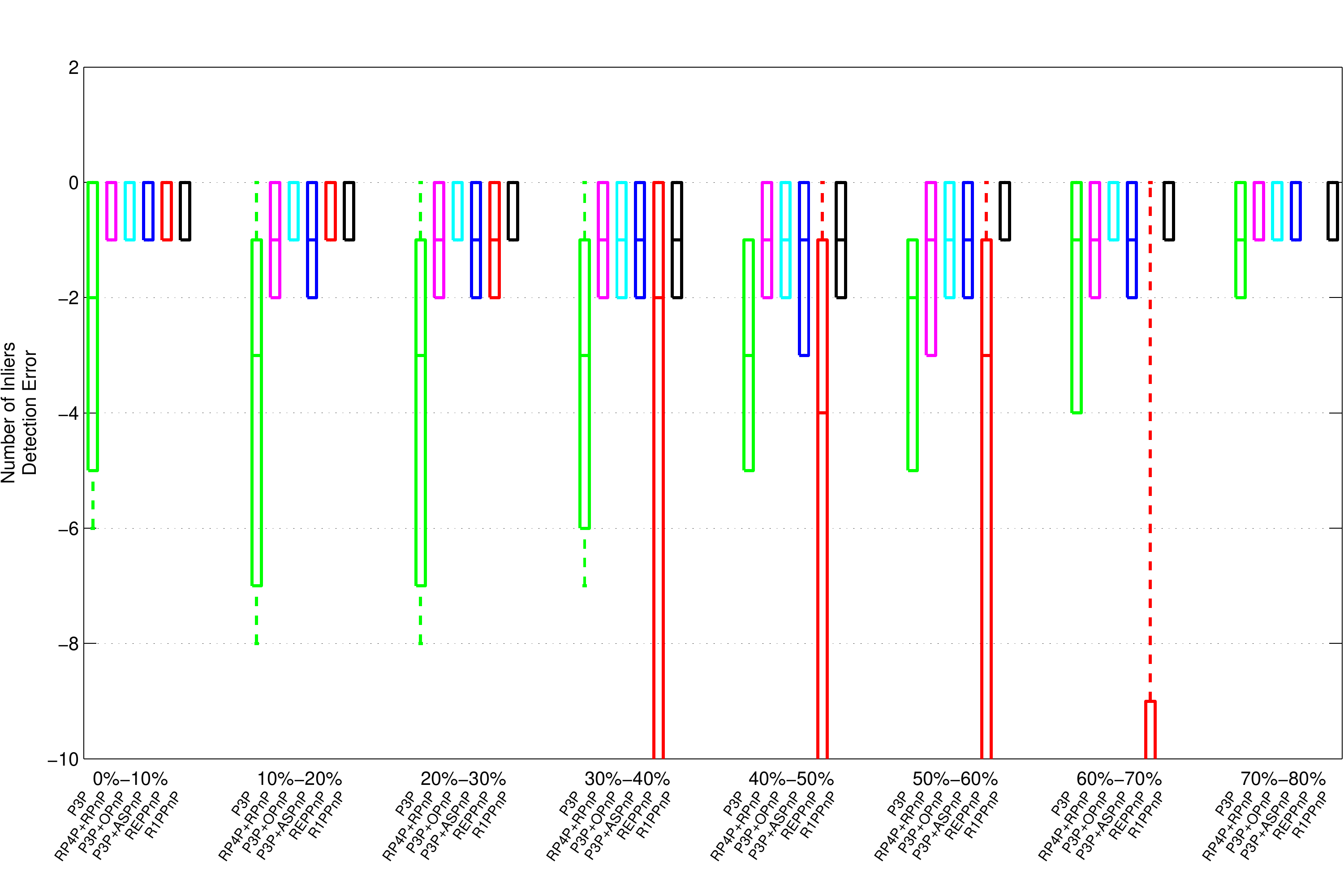}
  }
  \caption{Statistical results with real-world data. The $x$-axis is ranges of the percentage of outliers. (a) Rotation error. (b) Translation error. (c) Runtime. (d) The number of detected inliers compared with the maximum, zero suggests this method finds the most inliers.}
\label{fig_realworld}
\end{figure*}

The main advantage of R1PP$n$P is that it is capable of handling a large percentage of outliers with a much faster speed than conventional methods. For demonstrating this, we introduced the following RANSAC-based P$n$P methods for comparison: (RANSAC+P3P \cite{kneip2011novel}); (RANSAC + RP4P + RP$n$P \cite{li2012robust}); (RANSAC + P3P \cite{kneip2011novel} + ASP$n$P \cite{zheng2013aspnp}); and (RANSAC + P3P \cite{kneip2011novel} + OP$n$P \cite{zheng2013revisiting}). According to evaluations in outlier-free situations, OP$n$P is the most accurate P$n$P method and ASP$n$P and RP$n$P are fast. We selected these methods as the final refinement step to fully demonstrate the performance of RANSAC+refinement-like methods. Another important method is REPP$n$P \cite{ferraz2014very}, which is the state-of-the-art P$n$P algorithm that addresses outliers.

The experiments were conducted as follows. $N_{\rm{inlier}}=100$ correct matches (inliers) between 3D object points and 2D image points were generated. $N_{\rm{outlier}}$ mismatches (outliers) were generated by randomly corresponding 3D and 2D points. The true percentage of outliers is $p_{\rm{outlier}} = N_{\rm{outlier}} / (N_{\rm{inlier}}+N_{\rm{outlier}})$. Gaussian image noise with a standard deviation of $\sigma=5$ pixels was added. For R1PP$n$P and other RANSAC-based methods, the reprojection error threshold to distinguish inliers and outliers was $H = 10$ pixels.

As shown in Fig. \ref{fig_outliers}, REPP$n$P began to fail when the percentage of outliers was larger than $50\%$ with ordinary 3D cases, and only $5\%$ with quasi-singular cases. R1PP$n$P and RANSAC-based methods were capable of handling situations with a large percentage of outliers. R1PP$n$P was more accurate than RANSAC-based methods for both rotation and translation estimation. Compared to other RANSAC-based methods, R1PP$n$P was much faster, especially when the percentage of outliers was large.

\subsection{Real-world Image Data}

\begin{figure*} [htp]
\vspace{0.0cm}
\centering
  \includegraphics[width=.95\textwidth]{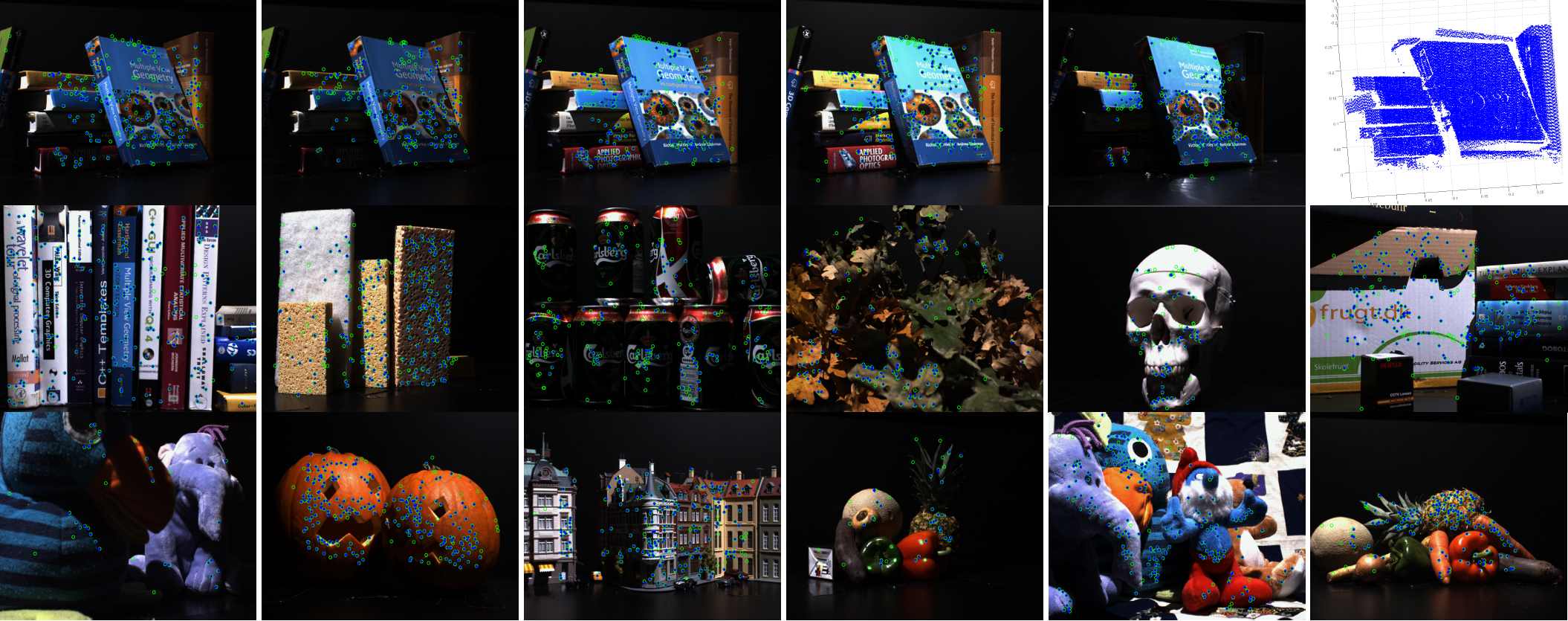} 
  \caption{Examples of images and R1P$n$P reprojection results. Green circles are all SURF correspondences and blue stars are the reprojected inliers detected by R1PP$n$P. First row: images with different illumination situations and the 3D point cloud. Second-third row: different data sets.}
\label{fig_realworldexamples}
\end{figure*}

Our real-world experiments were conducted on the DTU robot image data \footnote{http://roboimagedata.imm.dtu.dk/.} \cite{aanaes2012interesting}, which provides images and the related 3D point cloud obtained by structured light scan. The true values of rotations and translations are known. Images have a resolution of $800\times600$. Datasets numbered 1 to 30 were used. In each dataset, images were captured under 19 different illumination situations and from 119 camera positions. We selected 10 out of 19 illumination situations. Hence, a total of $30\times10\times119 = 35700$ images were included in this evaluation. Following the instruction, for each dataset and illumination situation, we used the image numbered 25 as the reference image and performed SURF matching \cite{bay2006surf} between the reference image and other images. The inliers threshold was $H = 5$ pixels for all methods. With each image, we ran all algorithms 5 times and used the average value for the subsequent statistics.

As shown in Fig.\ref{fig_realworldexamples}, the total number of correspondences and the percentage of outliers varied with objects, illumination situations and camera poses. Although clear comparisons require that only one factor is different, this kind of variable-controlling is difficult for P$n$P evaluation on real-world data because SURF matching results are unpredictable. In experiments we found that the performance of P$n$P algorithms were mainly affected by the percentage of outliers, rather than the total number of correspondences. Therefore in this section, we report the evaluation results by comparing the statistical results of P$n$P methods at each percentage of outliers range. Because the true number of inliers was unknown, for each image, algorithms detected inliers and we considered the maximum number of inliers as the ground truth.

As shown in Fig. \ref{fig_realworld}(a), as the percentage of outliers increased, the runtime of R1PP$n$P did not grow significantly compared with conventional RANSAC+P3P or P4P methods. When $p_{\rm{outliers}} < 30\%$, R1PP$n$P was slower than pure RANSAC+P3P, but was much more accurate as shown in Fig. \ref{fig_realworld}(a)(b). To improve accuracy, RANSAC+P3P needs other P$n$P methods, such as OP$n$P or ASP$n$P, as the final refinement step. Compared with other refinement P$n$P methods, R1PP$n$P was slightly less accurate than OP$n$P, which was the most accurate P$n$P method according to both synthetic and real-world experiments, but much faster even when the percentage of outliers was small.

 \begin{figure} [htp]
\vspace{0.0cm}
\centering
 \includegraphics[width=.45\textwidth]{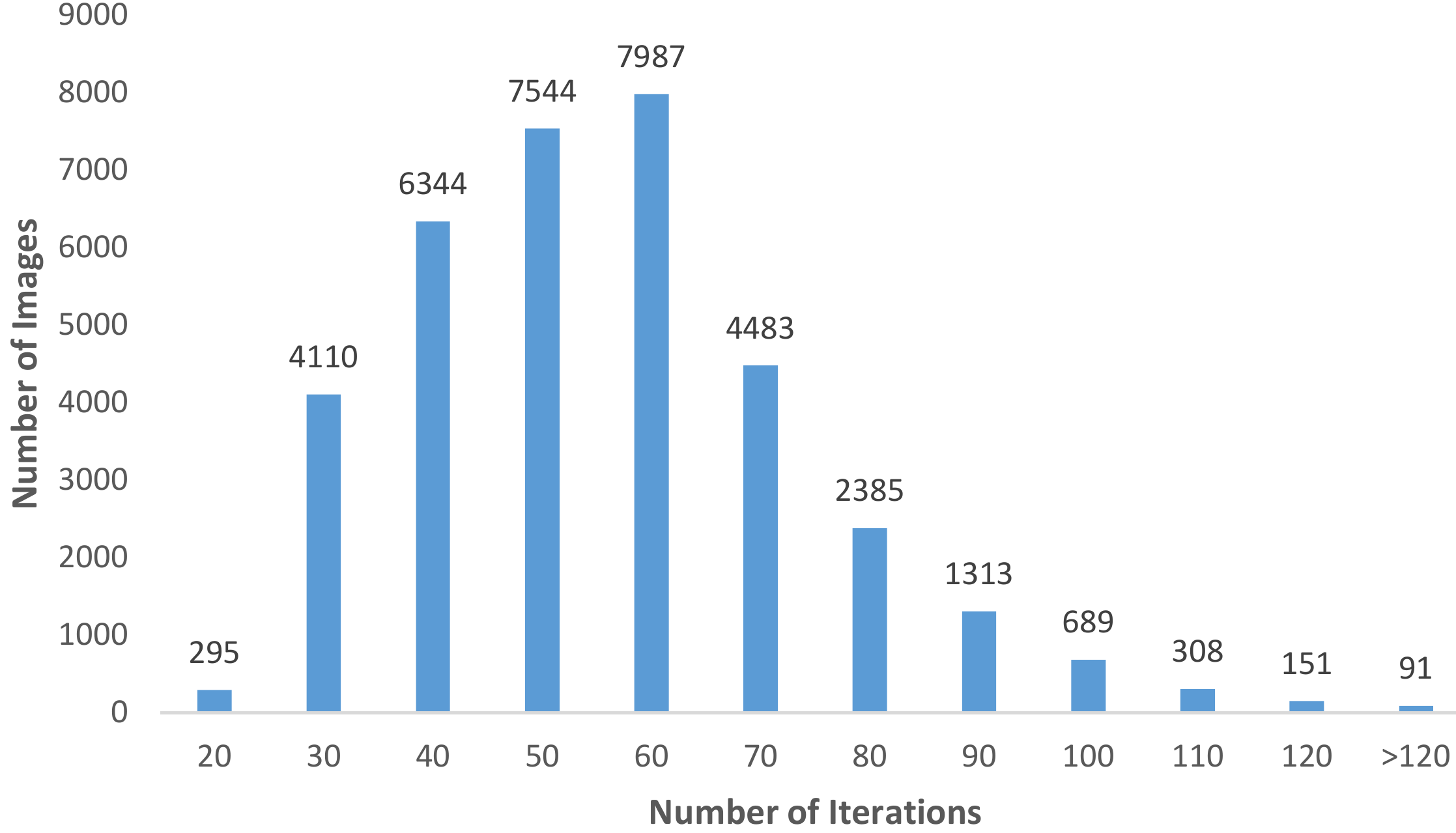}
\caption{Histogram of the number of iterations of R1PP$n$P in real-world experiments.}
\label{fig_realworld_iterations}
\end{figure}

Fig. \ref{fig_realworld_iterations} shows the histogram of the number of R1PP$n$P iterations  on all 35700 images. As shown in Fig. \ref{fig_algorithmstep}, the iteration number includes iterations with the re-weighting mechanism that obtained the best results in RANSAC trials, and the subsequent refinement iterations without re-weighting. The average number of required iterations is 51.3.

\section{Conclusions}

We present a fast and robust P$n$P solution named R1PP$n$P for tackling the outliers issue. We integrate a soft re-weighting method into an iterative P$n$P process to distinguish inliers and outliers, and employ the 1-point RANSAC scheme for selecting the control point. The number of trials is greatly reduced compared to conventional RANSAC+P3P or P4P methods; hence, it is much faster. Synthetic and real world experiments demonstrated its feasibility. Except for the good performance, another hidden advantage of R1PP$n$P is that its code implementation is relatively easy because all steps of R1PP$n$P involve only simple calculations. For example, its minima avoidance mechanism only requires to compute the determinant of the rotation matrix and to make $\lambda_{\rm{new}} = 1/\lambda_{\rm{old}}$. The most appropriate situations to replace conventional RANSAC+P3P methods with R1PP$n$P is when the percentage of outliers and/or the image white noise is large. R1PP$n$P is more appropriate for large point clouds because of its low time complexity and the requirement to try control points. Future works involve the development of its extension for planar cases, and applying it in the SLAM system to handle outliers when a new frame is encountered.

\section*{Acknowledgment}
This work is supported by NIH P41EB015898.

\ifCLASSOPTIONcaptionsoff
  \newpage
\fi

\bibliographystyle{IEEEtran}
\bibliography{IEEEabrv,bare_jrnl}

\end{document}